\documentclass[9pt,journal]{IEEEtran}

\usepackage{graphicx,epstopdf,subcaption}
\usepackage{multirow,multicol,booktabs,pgfplots}
\usepackage{amsmath}
\usepackage{amssymb,amsthm}
\usepackage{algorithm}
\usepackage[noend]{algpseudocode}

\usepackage{gensymb}
\usepackage{setspace}
\usepackage{cite}
\usepackage{times}
\usepackage[utf8]{inputenc}
\usepackage{color}
\pdfoutput=1

\begin{document}

\title{\LARGE \bf
Cooperative Planning for Fuel-constrained Aerial Vehicles and Ground-based Refueling Vehicles for Large-Scale Coverage
}

\author{Parikshit Maini$^{\$}$\thanks{$^{\$}$ Graduate Student, Dept. of CSE, IIIT-Delhi, India.},\;
Kaarthik Sundar$^{\dagger}$\thanks{$^{\dagger}$ Post Doctoral Researcher, CNLS, Los Alamos National Laboratory, USA.},\;
Sivakumar Rathinam$^{\ddagger}$\thanks{$^{\ddagger}$ Associate Professor, Dept. of Mechanical Engineering, Texas A\&M University, USA.},\;
PB Sujit$^{*}$\thanks{$^{*}$ Lecturer at Loughbrough University, UK and Assistant Professor at IIIT-Delhi, India.}\;
}

\maketitle
\begin{abstract}
	Low cost Unmanned Aerial Vehicles (UAVs) need multiple refuels to accomplish large area coverage. The number of refueling stations and their placement plays a vital role in determining coverage efficiency. In this paper, we propose the use of a ground-based refueling vehicle (RV) to increase the operational range of a UAV in both spatial and temporal domains. Determining optimal routes for the UAV and RV, and selecting optimized locations for refueling to aid in minimizing coverage time is a challenging problem due to different vehicle speeds, coupling between refueling location placement, and the coverage area at each location. We develop a two-stage strategy for coupled route planning for UAV and RV to perform a coverage mission. The first stage computes a minimal set of refueling sites that permit a feasible UAV route. In the second stage, multiple Mixed-Integer Linear Programming (MILP) formulations are developed to plan optimal routes for the UAV and the refueling vehicle taking into account the feasible set of refueling sites generated in stage one. The performance of different formulations is compared empirically. In addition, computationally efficient heuristics are developed to solve the routing problem. Extensive simulations 
 are conducted to corroborate the effectiveness of proposed approaches.
\end{abstract}

\begin{IEEEkeywords}
UAV; cooperative planning; route planning; fuel constraints; large scale coverage; aerial mapping
\end{IEEEkeywords}
\section{Introduction}
\label{sec:intro}

Mapping and surveillance are important exercises undertaken in a variety of applications like remote sensing \cite{review2014uav},  biodiversity surveys \cite{surveyForWildlife, sujit2007forestMonitoring} and Intelligence, Surveillance, and Reconnaissance (ISR) \cite{isr2005} missions. Many of these applications involve surveys on a massive scale, both in terms of size and manpower. Manual operations can be strenuous and time consuming, ranging anywhere from a few days to weeks \cite{physicalSurveysExpensive}. Furthermore, some regions may not be reachable by land due to terrain restrictions, leading to incomplete surveys. The use of manned aircrafts is possible but it is expensive and provides limited range with regards to imagery due to altitude restrictions \cite{surveyForWildlife}. Also, the noise generated by low flying manned aircrafts creates significant disturbance to the surroundings. Hence, small UAVs with on-board cameras have emerged as an economical alternative \cite{UAVsurvey2008, UAVsurvey2013BatteryProb} for mapping applications \cite{sujit2007forestMonitoring,isr2005}. They are capable of low-altitude, close-range imagery and generate less noise due to electric propulsion; additionally their hovering capability (in case of rotary-wing type aircraft) makes them suitable for stealth operations. 

\begin{figure}
	\centering
	\includegraphics[scale = 0.41]{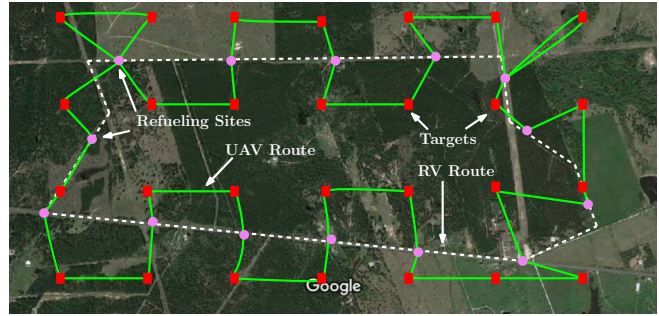}
	\caption{A sample scenario of a coverage application using a UAV. The UAV must visit all targets (red squares) to complete the mission. Due to limited fuel capacity, it needs to refuel to visit all targets. It may rendezvous with a RV, constrained to travel on the road-network, to refuel as needed. The figure depicts possible routes for the UAV and RV.}\label{fig:solution}	
\end{figure}

Even though there are several advantages of using a low-cost UAV, there also exist some limitations due to it's small size and payload. A small take-off weight combined with sensor-payload, constraints the fuel capacity of the UAV. 
Thus, for any reasonable-sized mission, it is not feasible for the UAV to visit all points of interest or \emph{targets} in a single sortie. The UAV may begin with maximum fuel, visit a set of targets, refuel at a refueling station and then resume the mission. It may need to refuel multiple times in order to visit all targets. Current state-of-the-art makes use of stationary refueling station(s) to refuel the UAV \cite{sundar2014TASE,kannon2014AFIT}. The placement of refueling stations affect coverage area size and incurs additional cost for each refueling station. To the best of our knowledge, there is no work in the literature that explores placement strategies for refueling stations to improve area coverage. This work investigates the use of a ground-based refueling vehicle (RV) to increase the operational range of a UAV in space and time. Figure \ref{fig:solution} illustrates a scenario for a coverage application. The UAV must visit all the targets (red squares) to complete the coverage mission. Limited fuel capacity makes the it incapable of visiting all targets in a single flight. The UAV must rendezvous with the RV, restricted to traverse on the road-network, to refuel. This admits placement strategies for UAV-RV rendezvous locations to improve coverage efficiency.

The success and cost of the mission, in terms of fuel consumed by the UAV,  depends on the identification of an appropriate set of rendezvous locations for the UAV and the RV, referred to as \emph{refueling sites}. The placement of refueling sites must ensure reachability of the targets and address fuel limitations of the UAV. In addition, as the RV is restricted to traverse on the road network, the refueling sites are restricted to be located on the road network. Furthermore, consecutive refueling sites must lie within road-distance that the RV can travel in one flight time of the UAV, to ensure timely refueling. The speed differential of the two vehicles and the fuel and terrain restrictions described above, build a strong coupling in the UAV and RV routes. In this scenario, the following fuel-constrained UAV routing problem with a mobile refueling station (FCURP-MRS) arises naturally: \\

\noindent \emph{Given a set of targets, a road network, a UAV and a RV; plan routes for the UAV and the RV such that each target is visited by the UAV, UAV and RV rendezvous at suitable locations on the road network to refuel as needed, UAV never runs out of fuel, and total distance traveled by the UAV is a minimum.} 

\noindent This work develops offline planning techniques to compute solutions to the problem and does not assume communication between the two vehicles. A two-stage strategy is designed for joint route planning for the UAV and RV to complete the mapping operation and rendezvous as needed. Contributions of this paper are the following:\\
\begin{itemize}
\item A two-stage approach to solve FCURP-MRS; the first stage computes a set of refueling sites on the road network while accounting for fuel constraints of the UAV and the speed differential of the two vehicles. The second stage solves the joint routing problem for the UAV and RV with refueling constraints and terrain restrictions.
\item MILP formulations are presented to solve the coupled UAV-RV routing problem, based on existing paradigms for fuel constrained routing problems. A branch-and-cut implementation framework to solve these MILP formulations is also described.
\item A computationally efficient heuristic to generate feasible solutions is detailed.
\item Performance of all approaches is corroborated using extensive simulations. Different approaches are benchmarked against baseline solutions based on solution quality and computation time, and trends are identified.
\end{itemize}

\section{Related work} \label{sec:literature}

FCURP-MRS is a generalization of the multiple vehicle, multiple depot, fuel constrained vehicle routing problem that is shown to be NP-hard \cite{sundar2015formulations}, and hence, is also NP-hard. The notion of a mobile refueling station creates a strong coupling between the routes for the UAV and RV, and makes the problem challenging. Current literature on the problem is scarce and there does not exist any method to compute the optimal solution. This section gives an overview of work done on related problems and solution techniques used therein. For finer details of these works, the reader is referred to the individual papers.

Khuller et. al \cite{khullerGasStation} were the first to propose the fuel-constrained routing problem in the context of a ground vehicle for a given set of stationary refueling stations; they developed a constant factor approximation algorithm for the problem. Fuel-constrained routing for a UAV, assuming that location of refueling stations are fixed and known a priori, has been addressed in \cite{sundar2014TASE} and \cite{kannon2014AFIT}. They develop MILP-based approaches and heuristics to determine UAV routes. Authors in \cite{sundar2016exact} and \cite{sundar2015formulations} present alternate formulations for the multiple vehicle, multiple depot variants of the problem. Heuristics for the multiple vehicle versions have been developed in \cite{Levy2014}. The problem addressed in this paper, FCURP-MRS, differs from aforementioned work in the sense that the refueling station is mobile.

Apart from UAVs, investigations with regard to routing of ground vehicles have also garnered considerable interest in the literature. Route planning for ground vehicles, such that the vehicle never runs out of fuel is a related problem. Khuller et. al. \cite{khullerGasStation} consider the problem of path planning in the presence of refueling stations with varying fuel prices and develop a dynamic programming solution to compute an optimal solution for $s-t$ shortest path version of the problem. Erdogan and Miller-Hooks \cite{erdougan2012green} consider the vehicle routing problem for alternative fuel vehicles. They develop construction heuristics and validate performance using numerical experiments. The complementary problem of optimal placement of recharging stations for electric vehicles to guarantee energy supply on any shortest path has been studied in \cite{funke2015placement}. Mathew et. al. \cite{smith2015TRO} use mobile refueling stations to refuel the UAVs in persistent missions. They assume a priori knowledge of UAV trajectories and develop strategies on the use of multiple ground-based refueling vehicles to refuel the UAVs. The objective of their problem is to determine a schedule for refueling for the UAVs and to compute routes for the ground vehicles to meet the schedule. This work addresses the more general problem on joint route planning problem both UAV and ground vehicle, to extend the operational range of the UAV in both space and time. Authors in \cite{ICUASself,2Echelon17} develop construction heuristics for a similar problem, but do not benchmark their results.

To the best of our knowledge, the current state of the art on joint-route planning of a UAV and refueling ground vehicle is inadequate and there do not exist mathematical formulations or techniques to estimate the optimum. This work aims to bridge this gap in the literature by addressing the FCURP-MRS applied to a coverage application. The remainder of the paper is organized as follows: 
Section \ref{sec:appScenario} gives an overview of the problem scenario and the underlying assumptions. Section \ref{sec:siteSelection} presents the site selection algorithm that forms the first stage of the solution approach. Sections \ref{sec:routePlanning} and \ref{sec:heuristic} develop algorithms for the second of the solution approach. In particular, the Section \ref{sec:routePlanning} develops MILP formulations for the routing problem and presents a branch-and-cut algorithm to solve the MILP formulations and Section \ref{sec:heuristic} details a computationally efficient heuristic algorithm. Section \ref{sec:simResults} compares the various algorithms and presents simulation results. 
 Concluding remarks and future research directions are discussed in Section \ref{sec:concFut}.

\begin{figure*}
\begin{subfigure}{0.5\linewidth}
	\centering
	\includegraphics[scale = 0.41]{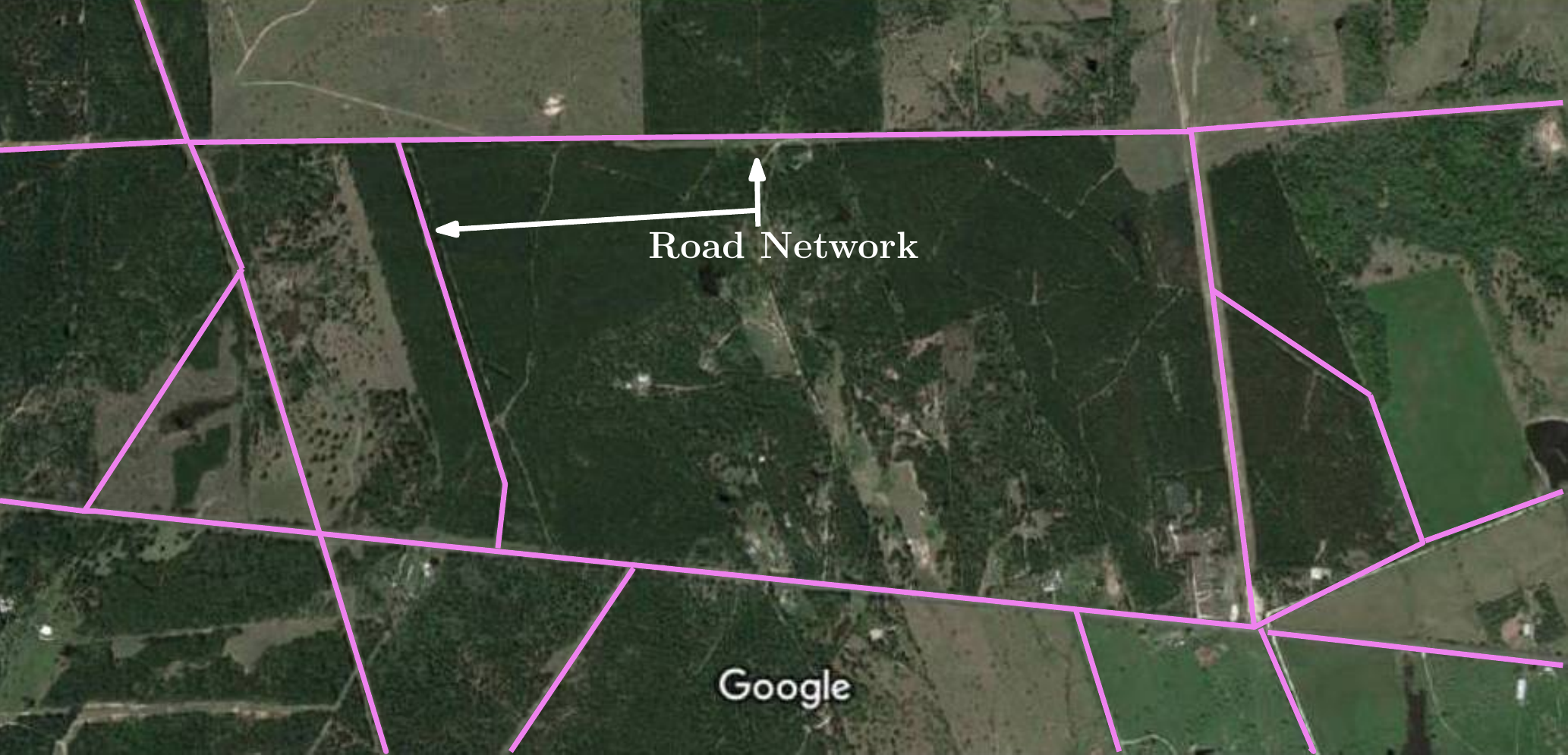}
	\caption{}\label{fig:scenario}	
\end{subfigure}
\begin{subfigure}{0.5\linewidth}
	\centering
	\includegraphics[scale = 0.41]{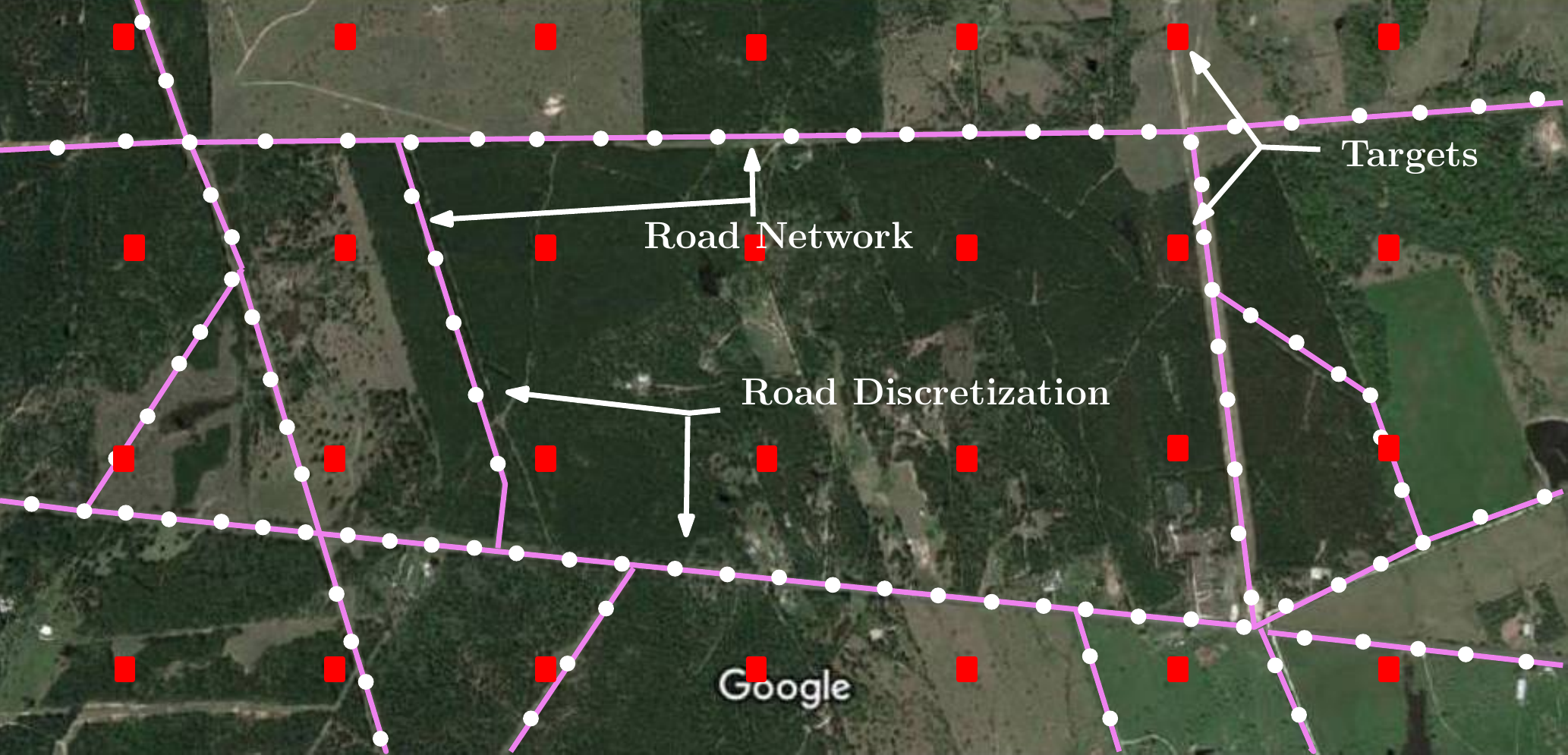}
	\caption{}\label{fig:discretizedRN}	
\end{subfigure}
\caption{(a). A characteristic environment with an interior road network. Refueling must happen along the road network as the RV is restricted to traverse on the road network. (b). The discretized locations on the road network that form the initial set of candidate refueling sites $\mathcal S$.}\label{fig:illustration}
\end{figure*}

\section{Application scenario}\label{sec:appScenario}

Consider an environment ($\mathcal E$) with an interior road network as shown in Fig. \ref{fig:illustration}(a), a UAV with a fixed down-facing camera on board and a RV that traverses on the given road network. To perform a mapping mission, $\mathcal E$ is discretized into cells of equal size. Size of a cell is equal to the camera foot print of the UAV at a given altitude. The grid centers are denoted as targets ($T$) in $\mathcal E$. UAV has limited fuel capacity and cannot visit all targets in a single flight. This necessitates refueling, possibly multiple times, to complete the mission. The UAV must rendezvous with the RV to refuel. A solution to the problem comprises of routes for both UAV and RV that satisfy respective constraints of the two vehicles and visits all targets in $T$.

\subsection{UAV Model} \label{subsec:uavModel}

UAV model considered is a multi-rotor or VTOL (Vertical Take-Off and Landing) type vehicle. Flying conditions are ideal and effects due to wind are neglected. It travels at a constant altitude and a constant air speed, $V_u$. The UAV autopilot uses autonomous landing techniques, \cite{gautam2014ICUAS} and references within, for instance vision-based methods \cite{gautam2017IFAC} to land on the RV. The refueling process on the RV may be automated using battery swap systems \cite{dock1, dock2} or performed manually by a human operator. 

Further, the following assumptions are made on the UAV model. Fuel consumed by the UAV to travel from location $i$ to $j$ is proportional to euclidean distance (in 2D) between $i$ and $j$ and is independent of flight maneuvers. For ease of exposition, the proportionality constant is assumed to be unity. Given constant speed for the UAV, distance traveled by the UAV is also proportional to flight duration. Hence, distance traveled, time of flight and fuel consumed are used interchangeably through the rest of the text, unless otherwise indicated. Let $U$ be the fuel capacity of the UAV available after compensating for a take-off and landing sequence; due to the previous assumption, $U$ is also the maximum distance the UAV can travel when starting at full fuel capacity. It may hover at the refueling site if the RV has not reached, but the sum of traveling time and hovering time must not exceed $U$. 

\subsection{Refueling Vehicle Model}\label{subsec:rvModel}
 RV travels with a constant speed, $V_r$, between refueling sites and remains stationary during rendezvous operations with the UAV. For the sake of calculation, let $t_{u} = U/V_u$ denote the maximum flight time of the UAV in one sortie, then $R = t_u V_r$ is the maximum distance traveled by the RV, on road, in one flight time of the UAV. The road-distance traveled by the RV between two points in the environment is at least as large as the euclidean distance between the two points. The RV has a sufficient supply of UAV batteries and never runs out of fuel itself.

Since the aim of this work is to plan routes for the UAV and RV, vehicle dynamics and kinematics are not considered. The constant speed assumption on both UAV and RV is non-binding and is used as an abstraction to keep the focus of the problem on route planning. For the UAV, it is used to quantify the maximum flight time. In the case of RV, it is only used to compute the value of $R$ that may also be computed by the knowledge of maximum speed of the RV. The infinite fuel supply to RV assumption is also not necessary and is used to convey the general idea that payload capacity of the RV is not restrictive and it may be readily refueled.

\section{Refueling Site Selection} \label{sec:siteSelection}

The refueling site selection problem is a precursor to route planning for both UAV and RV. A greedy algorithm (Algorithm \ref{algo:DCHS}) is presented to compute a set of refueling sites so that the joint optimization problem in the second stage is feasible.

To begin with, the road network is discretized at a resolution $\triangle << R$. The set of candidate refueling sites, denoted by $\mathcal S$, comprises of the discretized locations on the road network (Fig. \ref{fig:discretizedRN}). Typically, the cardinality of $\mathcal S$ is very large. For the sake of computational tractability, a small set of refueling sites that ensures the existence of a feasible solution to the joint UAV-RV routing problem, is required. Let $T$ denote the set of targets. For a subset, $S$, of candidate refueling sites, $\mathcal S$, to be deemed as a valid set of refueling sites, it must satisfy the following two conditions:
\begin{enumerate}
\item {\bf Coverage condition:} Each target $t \in T$ has a refueling site $s \in S$ such that the distance between $s$ and $t$ is at most $U/2$ units. This distance also corresponds to the fuel consumed by the UAV to travel from $s$ to $t$.
\item {\bf Connectedness condition:} Refueling sites in $S$ form a connected component in the graph $G_{r}$, where $G_{r}\equiv (S,E)$; an edge $(i,j)$ where $i,j \in S$ is contained in the set $E$ if and only if the distance between $i$ and $j$ via the road network (road distance) is at most $R$ units.
\end{enumerate}

The above conditions ensure that if there exists a solution to the site selection problem there always exist a feasible solution to the joint routing problem of a UAV and RV in the presence of refueling constraints for the UAV.

\begin{figure}
\centering
\includegraphics[scale = 0.41]{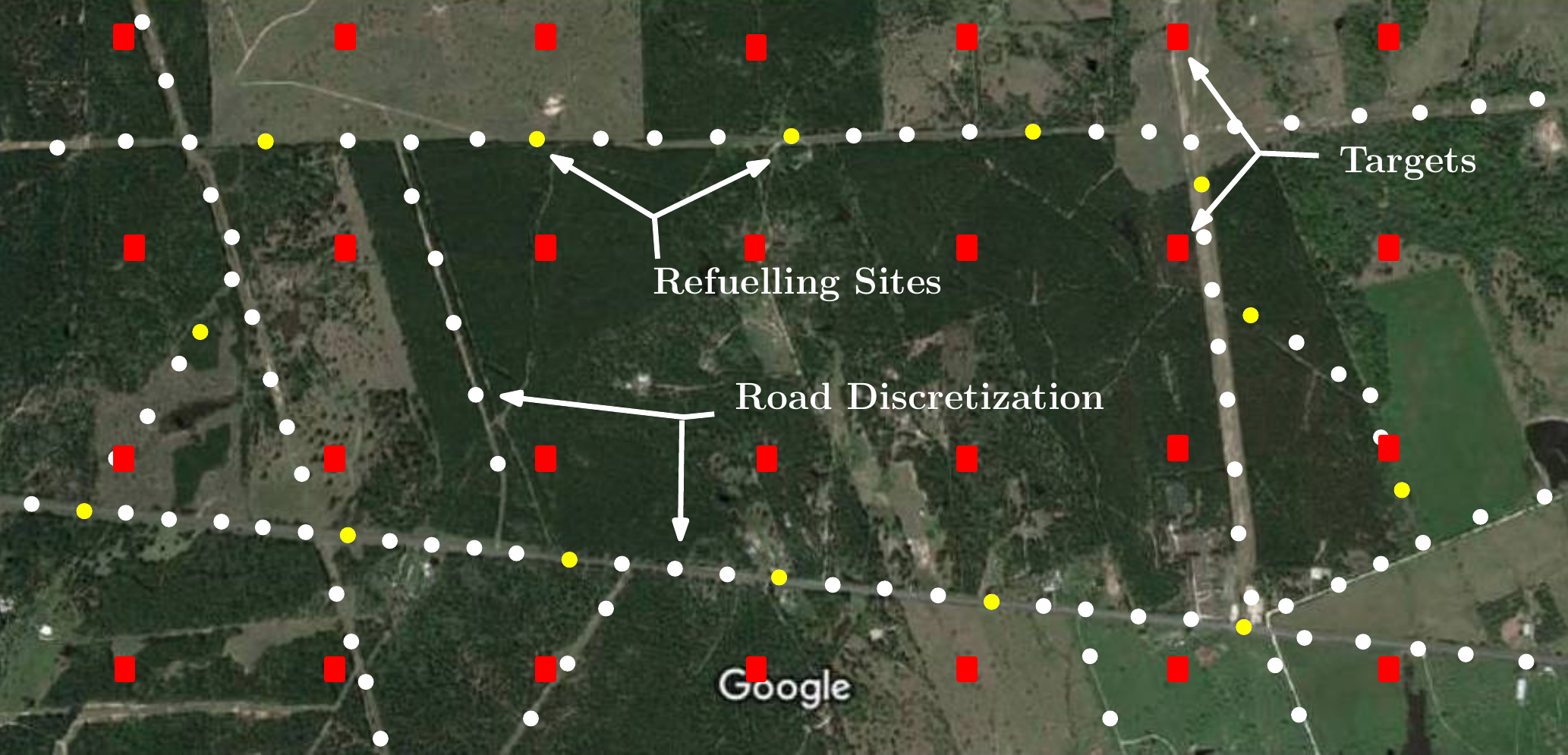}
\caption{The set of refueling sites obtained after the first stage using the algorithm described in Section \ref{sec:siteSelection}. This set of refueling sites ensure a feasible solution for the joint routing problem in the second stage.}
\label{fig:siteselection}
\end{figure}

The objective of the refueling site selection problem is to compute a subset $S \subseteq \mathcal S$ of low cardinality that satisfies \emph{coverage} and \emph{connectedness} conditions. To that end, let $H(s)$ and $N(s)$ be two sets associated with every candidate refueling site $s \in \mathcal{S}$ . The \emph{reachable set}, $H(s)$, comprises of all targets reachable from $s$, while the \emph{neighbor set}, $N(s)$, comprises of neighboring candidate refueling sites of $s$ within $R$ road distance. A target $t \in T$ is said to be reachable from $s$ if the distance between $t$ and $s$ is at most $U/2$. To compute such a minimum cardinality subset $S$, a greedy algorithm as detailed in Algorithm \ref{algo:DCHS} is used. The algorithm iteratively selects a refueling site $s \in \mathcal S$, that covers the maximum number of uncovered targets, and adds it to $S$. This process is repeated until all targets are reachable from $S$, i.e. $H(S) = T$, where $H(S) = \bigcup_{s\in S} H(s)$. An illustration of the solution obtained using the greedy refueling site selection algorithm (Algorithm \ref{algo:DCHS}) is shown in Fig. \ref{fig:siteselection}. The time complexity of the algorithm is $\mathcal O{(|T| |\mathcal S|)}$.

 \begin{algorithm}
 \caption{Refueling Site Selection}\label{algo:DCHS}
 \begin{algorithmic}[1]
 \Require $\mathcal{S}, T, H(s)$ and $N(s)$ 
\Statex $\backslash\backslash$ $H(S) = \bigcup_{s\in S} H(s) ~\forall s\in \mathcal S$
\Statex $\backslash\backslash$ $N(S) = \bigcup_{s\in S} N(s) ~\forall s\in \mathcal S$
 \State $s_{max}=\underset{s \in \mathcal S}{\rm argmax} ( | H(s) | )$ $\backslash\backslash$  mission starting point 
 \State $S = \{s_{max}\}$
 \State $T = T\setminus H(S)$
 \While{$T\neq \phi$}
 \State $s_{max}=\underset{s \in N(S)}{\rm argmax} ( |H(s)\setminus H(S) | ) $
 \State $S = S \bigcup \{s_{max}\}$
 \State $T = T\setminus H(s_{max})$
 \EndWhile
 \State \Return $S$
 \end{algorithmic}
 \end{algorithm}
 
 \section{Routing problem}\label{sec:routePlanning}

This section formalizes the fuel constrained UAV routing problem with mobile refueling station. The minimal set of refueling sites obtained from the refueling site selection algorithm in Sec. \ref{sec:siteSelection} is an input to the UAV-RV joint routing problem, defined as follows: Given a set of targets, $T$, and a set of refueling sites, $S$, plan tours for the UAV and the RV such that the UAV visits each target exactly once, UAV and RV rendezvous at refueling sites in $S$ to refuel, UAV never runs out of fuel and the total fuel consumed by the UAV is a minimum. The UAV and RV are initially stationed at a common refueling site and return to the same location after the mission.
 
Let $\mathcal P$ represent a feasible UAV route for the routing problem. Let $\mathcal{P}_i$ be the $i$\textsuperscript{th} subpath of $\mathcal{P}$, traversed by the UAV between two consecutive refueling sites, $s_j$ and $s_k$. Also, let $\mathcal{F}_i$ be the fuel consumed and $\tau_i$ be the set of targets visited by UAV on $\mathcal{P}_i$ and $r_{jk}$ be the road distance between $s_j$ and $s_k$. Then $\mathcal{P}$ satisfies the following constraints: (i) $\bigcup_{i} \tau_i = T$ \emph{i.e.}, all the targets are visited by the UAV, (ii) $\tau_i$ $\bigcap \tau_j = \emptyset, \forall i,j,~i\neq j$ \emph{i.e.}, each target is visited exactly once by the UAV, (iii) $\mathcal{F}_i \leq U, \forall i$ \emph{i.e.}, the UAV never runs out of fuel, and (iv) $r_{jk} \leq R, \forall i$ \emph{i.e,} the refueling sites $s_j$ and $s_k$ at the start and end of each subpath $\mathcal P_i$ are within road distance $R$ to ensure that the RV reaches $s_k$ before the UAV runs out of fuel. The objective of the problem is to minimize the total fuel consumed (proportional to distance traveled) by the UAV,
\begin{equation}
\min \sum_{i:\mathcal{P}_i \in \mathcal{P}} \mathcal{F}_i.
\end{equation}
The fuel constrained UAV routing problem with mobile refueling station (FCURP-MRS)  problem addressed in this work can be formulated in the Mixed Integer Linear Programming (MILP) framework. 
 The FCURP-MRS is formulated on a complete directed graph $G = (V,E)$, with vertex set $V = T \bigcup S$ and edge set $E$. The UAV and the RV are initially stationed at a refueling site $s_0 \in S$. Associated with the edge set are two weight functions: $f: E \rightarrow \mathbb{R}^{+}$, where $f_{ij}$ denotes the fuel consumed by the UAV when it travels along the directed edge $(i,j)$ and $r: (S\times S) \rightarrow \mathbb{R}^{+}$, that represents the road distance between two refueling sites. Let $N:S\rightarrow \wp(S)$, where $\wp(S)$ is the power set of $S$, denote a neighborhood function defined as $N(s_i) := \{s_j:  r_{ij} \leq R, s_j \in S\}$. MILP formulation can be developed using node-labeling or edge-labeling approach. Formulations based on both paradigms and a branch-and-cut based implementation to determine the optimal solution are detailed next.

\subsection{Node-based formulation} \label{subsec:node}
in this formulation decision variables on each vertex of the graph $G$ are used to formulate FCURP-MRS. The set of decision variables used in the formulation are defined as follows. $x_{ij}$, defined for each edge $(i,j) \in E$, are binary variables that represent whether or not the UAV traverses the edge $(i,j)$. Associated with each target $t_i \in T$ is a variable $u_i$ that represents the amount of fuel left in the UAV when it reaches $t_i$. $y_{ij}$ are binary decision variables, defined for every pair $<t_i,s_j>: t_i \in T$ and $s_j \in S$, and take value 1 if $s_j$ is the most recently visited refueling site when the UAV is at target $t_i$. Also, for any subset of vertices $P \subseteq V$, define $\delta^+(P) := \{(i,j): (i,j) \in E, i\in P, j \notin P\}$. The objective of FCUPR-MRS is to minimize the fuel consumed by the UAV during the mission, given as:\\

\noindent \emph{Objective}:
\begin{flalign}
&\mathcal F_1:~ \min \sum \limits_{i \in V} \sum \limits_{j \in V} f_{ij}x_{ij}. & \label{eq:obj} 
\end{flalign}

\noindent Constraints of the problem are expressed as follows,

\noindent \emph{Degree constraints}:
\begin{flalign}
& \sum_{i\in V\setminus{\{j\}}} x_{ij} = \sum_{i \in V\setminus{\{j\}}} x_{ji}, \quad \forall j \in V \text{ and}\label{constr:tourConst} &\\
& \sum_{i\in V\setminus{\{t\}}} x_{it} = 1, \quad \forall t \in T.\label{constr:visitTarget} &
\end{flalign}
The degree constraints in Eq. \eqref{constr:tourConst} enforce in-degree to be equal to out-degree $\forall j \in V$. Constraint \eqref{constr:visitTarget} ensures that each target is visited exactly once by the UAV. By not limiting the in/out-degree for refueling sites, UAV is permitted to make multiple visits to the refueling sites.\\

\noindent \emph{Sub-tour elimination constraints}:
\begin{flalign}
&\sum_{(i,j) \in \delta^+(P)} x_{ij} \geq 1, \quad\forall P \subseteq V\setminus \{s_0\}, P \cap T \neq \phi. &\label{constr:cutSet}
\end{flalign}
Constraint set \eqref{constr:cutSet} eliminates sub-tours in the UAV route by enforcing a path to exist from the initial refueling site to every target in the set $T$. It may be observed that the number of such constraints in the formulation is exponential; a dynamic cut-generation procedure is detailed in Section \ref{sec:bnc} to add these constraints into the problem as needed, without having to enumerate all of them.\\

\noindent \emph{Fuel constraints}:
\begin{flalign}
& u_{t} - u_{j} + f_{jt} \leq M(1-x_{jt}), \quad\forall t,j \in T, \label{constr:t2tFuel1} &\\
& u_{t} - u_{j} + f_{jt} \geq -M(1-x_{jt}), \quad\forall t,j \in T, \label{constr:t2tFuel2} &\\
& u_{t} - U + f_{kt} \leq M(1-x_{kt}), \quad\forall t \in T, \forall k \in S, \label{constr:d2tFuel1} &\\
& u_{t} - U + f_{kt} \geq -M(1-x_{kt}), \quad\forall t \in T, \forall k \in S, \label{constr:d2tFuel2} &\\
& -u_{t} + f_{tk} \leq  M(1-x_{tk}), \quad\forall t \in T, \forall k \in S, \label{constr:t2dFuel} &\\
&f_{ij}\cdot x_{ij} \leq U \quad \forall i,j \in S, \label{constr:d2d}&\\
& \sum_{i \in V} \sum_{j \in V} f_{ij}x_{ij} \leq U \sum_{k \in S} \sum_{i \in V} x_{ki}. \label{constr:totalFuel} &
\end{flalign}
The fuel constraints \eqref{constr:t2tFuel1}--\eqref{constr:totalFuel} ensure that the UAV does not run out of fuel as it traverses its route. In particular, constraints \eqref{constr:t2tFuel1} and \eqref{constr:t2tFuel2} ensure fuel conservation when the UAV travels between two targets. Constraints \eqref{constr:d2tFuel1}--\eqref{constr:t2dFuel} enforce similar restrictions on the UAV when it travels between a refueling site and a target. In all of these constraints, $M$ represents a large constant $M = U + \max_{(i,j)} ~ f_{ij}$. It may be noted that UAV can reach a refueling site with some fuel left in the vehicle. Constraint \eqref{constr:d2d} restricts direct paths between refueling sites to exist only between sites at most $U$ distance away. Constraint \eqref{constr:totalFuel} states that the total fuel consumed by the UAV must be at most equal to $U$ times the total number of refueling visits.\\

\noindent \emph{Refueling site constraints}:
\begin{flalign}
& y_{ts} - x_{st} \geq 0, \quad\forall t \in T, s \in S,\label{constr:directSource} &\\
& y_{t_2s} - y_{t_1s} \leq (1-x_{t_1t_2}), \quad\forall s \in S, \forall t_1, t_2 \in T\label{constr:indirectSource1}, &\\
& y_{t_2s} - y_{t_1s} \geq -(1-x_{t_1t_2}), \quad \forall s \in S, \forall t_1, t_2 \in T,\label{constr:indirectSource2} &\\
& \sum \limits_{\substack{{k \in S\setminus{N(s)}}}} x_{tk} \leq (1- y_{ts}), \quad\forall t \in T, s \in S,\label{constr:neighborRestrict1} &\\
&\sum \limits_{\substack{{k \in S\setminus{N(s)}}}} x_{sk} =  0, \quad \forall s \in S, \label{constr:neighborRestrict2}&\\
&\sum \limits_{s \in S} y_{ts}  = 1, \quad\forall t \in T.\label{constr:oneMRVdepot}&
\end{flalign}
Constraints \eqref{constr:directSource}--\eqref{constr:oneMRVdepot} limit the road distance between consecutive refueling sites to be at most $R$. Constraints \eqref{constr:directSource}--\eqref{constr:indirectSource2} set the value of $y_{ts}$ variables to appropriate value using UAV route decision variables.\\

\noindent \emph{Variable restrictions}:
\begin{flalign}
& x_{ij} \in \{0,1\}, \quad \forall (i,j) \in E, {\rm either}~i~{\rm or}~j \in T \label{constr:xDomain} &\\
& u_i \in [0,U], \quad \forall i \in T,&\label{constr:uDomain}\\
& y_{ts} \in {\{0,1\}}, \quad\forall t \in T, s \in S.\label{constr:yDomain} &
\end{flalign}
Finally, constraints \eqref{constr:xDomain}, \eqref{constr:uDomain}, and \eqref{constr:yDomain} enforce domain restrictions and bounds on $x_{tk}$, $u_i$, and $y_{ts}$ decision variables.

\subsection{Edge-based formulation} \label{subsec:edge}
In this formulation decision variables on each edge of $G$ are used to formulate fuel constraints for the UAV. Binary decision variables $x_{ij}$ and $y_{ij}$ introduced for node-based formulation in the previous section are retained in the edge-based formulation while continous decision variables $u_i$ are replaced with $z_{ij}$ variables, defined for each $(i,j) \in E$, to formulate fuel constraints. The decision variable $z_{ij}$, represents the amount of fuel used by the UAV to reach the $j^{th}$ vertex from a (most recent) refueling site when traveling along the incoming edge $(i,j)$. The incoming edge included in the UAV tour is uniquely defined for each target. Hence, the differentiating factor between node-based and the edge-based formulations for FCURP-MRS is in the way fuel constraints for the UAV are formulated.\\

\noindent \emph{Fuel constraints}:
\begin{flalign}
&\sum\limits_{i\in V}z_{ti} - \sum\limits_{i\in V}z_{it} = \sum\limits_{i\in V} f_{ti} x_{ti}, \quad\forall t\in T,&\label{constr:zDef}\\
&z_{ki} = f_{ki} x_{ki}, \quad\forall k\in S, \forall i\in V,&\label{constr:zDepot}\\
&0\leq z_{ij}\leq U x_{ij}, \quad\forall i,j \in V,&\label{constr:zRange}\\
&z_{ij} \in \mathbb{R}^+, \quad\forall i,j \in V.&\label{constr:zDomain}
\end{flalign}
The new formulation $\mathcal F_2$ replaces the fuel constraints \eqref{constr:t2tFuel1}-\eqref{constr:totalFuel} and variable restrictions on the $u_i$ variables  \eqref{constr:uDomain}, with constraints \eqref{constr:zDef}-\eqref{constr:zRange} and \eqref{constr:zDomain}, respectively. Constraint \eqref{constr:zDef} conserves the fuel at each target and \eqref{constr:zDepot} addresses the terminal cases. Constraints \eqref{constr:zRange} and \eqref{constr:zDomain} specify the upper and lower bounds on $z_{ij}$ variables.

\subsection{Refueling Vehicle Route}
Route for the ground based refueling vehicle is computed from the solution to the MILP formulation. The formulation ensures that consecutive visits to refueling sites on the UAV tour are within $R$ road distance. The tour comprising refueling site visits in the sequence as they occur on the UAV tour is a valid tour for the RV and ensures feasibility of the UAV tour. The tour so computed, ensures that the RV, when traveling at the constant speed $V_R$, always reaches the refueling site before the UAV runs out of fuel. The sum of traveling time and hovering time for the UAV is thus always constrained to be within $U$, allowing the UAV to land on the RV.

\subsection{Analysis}\label{sec:f1Anal}
Let $m = |T|$, $p = |S|$ and $n = |V|$. The node-based formulation uses $n^2$ $x_{ij}$ variables, $m$ $u_i$ variables, and $(m p) ~y_{ts}$ variables. Total number of variables is $n^2 + (p+1) m \sim \mathcal{O}(n^2)$. The number of constraints may be computed as : $(m+n)$ degree constraints, $2^n$ sub-tour elimination constraints, $(2m^2 + 3 m p+m^2 +1)$ fuel constraints, and $(2m p + 2m^2 p + m)$ refueling site constraints in addition to variable restrictions. Hence, the total number of constraints is $(1+2m+n+ 5m p+2m^2+2m^2 p+2^n) \sim \mathcal{O}(2^n)$. A similar analysis can also be performed for the edge-based formulation; the number of decision variables and the number of constraints is of the order $\mathcal{O}(n^2)$ and $\mathcal{O}(2^n)$ respectively. Since the number of constraints is exponential, it is computationally intractable to enumerate all constraints.

 \begin{algorithm}
 \caption{Separation Algorithm}\label{algo:sepAlgo}
 \begin{algorithmic}[1]
\State Build graph $G$(directed) $\equiv (V,E)$\label{sepAlgo:buildGraph}
\State Add edge $(i,j)$ to $E$, for each $x_{ij} = 1$\label{sepAlgo:populateGraph}
\State $\mathcal P =$  strongly connected components in $G$\label{sepAlgo:SCC}
\ForAll{$P \in \mathcal P$}\label{sepAlgo:forLOOPscc}
\If{$(|P|>1) \&\& ( P\subseteq \mathcal V\setminus \{s_0\}) \&\& (P \cap T \neq \phi)$}\label{sepAlgo:findSubTour}
\State Add violated constraint (Eq. \eqref{constr:sepAlgo}) \label{sepAlgo:addCosntraint}
\EndIf
\EndFor
 \end{algorithmic}
 \end{algorithm}

\subsection{Branch-and-cut algorithm}\label{sec:bnc}

To optimally solve the two formulations presented in the Sec. \ref{subsec:node} and \ref{subsec:edge}, a branch-and-cut implementation is employed. Due to the presence of  sub-tour elimination constraints \eqref{constr:cutSet}, it is not computationally efficient to enumerate all constraints in Eq. \eqref{constr:cutSet} and provide them to an MILP solver. To address this issue sub-tour elimination constraints in the formulation are relaxed, and whenever the solver obtains an integer solution feasible to this relaxed problem, a check is made to find if any of the relaxed constraints are violated by the feasible solution. If so, the violated constraints are added to the formulation and the problem is given back to the solver. This process of adding constraints to the problem sequentially has been observed to be computationally efficient for the many variants of the traveling salesman problem \cite{Grotschel1985} and also the fuel constrained vehicle routing problems \cite{sundar2016exact}. The algorithms used to identify violated constraints, also called \emph{valid inequalities}, given an integer feasible solution are referred to as separation algorithms. Algorithm \ref{algo:sepAlgo} presents the pseudocode for a separation algorithm used to dynamically identify violated constraints \eqref{constr:cutSet} given an integer feasible solution for FCURP-MRS. The algorithm computes the strongly connected components (line \ref{sepAlgo:SCC}) in the graph defined by the integer feasible solution. Each component $P$ that satisfies the condition $P \subseteq V\setminus \{s_0\}$, and $P \cap T \neq \phi$ (line \ref{sepAlgo:findSubTour}) violates the corresponding constraints as given in Eq. \eqref{constr:sepAlgo}. The violated constraints are then added to the formulation and the solver is allowed to optimize the problem with the new constraints.

\begin{flalign}
&\sum_{(i,j) \in \delta^+(P)} x_{ij} \geq 1, \quad P \cap T \neq \phi. &\label{constr:sepAlgo}
\end{flalign}

\section{Heuristics} \label{sec:heuristic}

Even though the MILP formulation and the branch-and-cut algorithm presented in the previous section compute an optimal tour for FCURP-MRS, expectedly the computation time grows exponentially as the number of targets in the environment increases. In such cases, it is useful to find a \emph{good} solution, quickly. This section develops a fast heuristic to compute feasible solutions for FCURP-MRS.

The heuristic works as follows. It compute a TSP (Traveling Salesman Problem) tour of the targets and the initial refueling site using Lin-Kernighan-Helsgaun heuristic \cite{lkh}. This tour corresponds to the UAV route. If the UAV does not run out of fuel as it traverses the tour then a feasible solution to FCURP-MRS has been found, else a repairing algorithm (Algorithm \ref{algo:heuristicRepair}) is used to convert the tour in to a feasible solution for FCURP-MRS that satisfies fuel and road distance constraints.

 \begin{algorithm}
 \caption{Repair Algorithm}\label{algo:heuristicRepair}
 \begin{algorithmic}[1]
\State current vertex = start of tour
\While{current vertex $\neq$ end of tour} \label{line:whileFuel}
\State Check for fuel capacity violation
\If{found a fuel capacity violation}
\State break
\Else
\State Go to next vertex\label{line:endwhileFuel}
\EndIf
\EndWhile
\While{feasible path not found}
\State compute \emph{indirectPath} to fix the violation\label{line:callIndirectPath}
\If{feasible path found}
\State break
\Else
\State back track\label{line:backTrack}
\EndIf
\EndWhile
\State recursive call to Repair Algorithm on subsequence starting from last depot on the feasible indirect path\label{line:recursiveCall}
 \end{algorithmic}
 \end{algorithm}

\begin{figure}
	\centering
	\includegraphics[width = \linewidth]{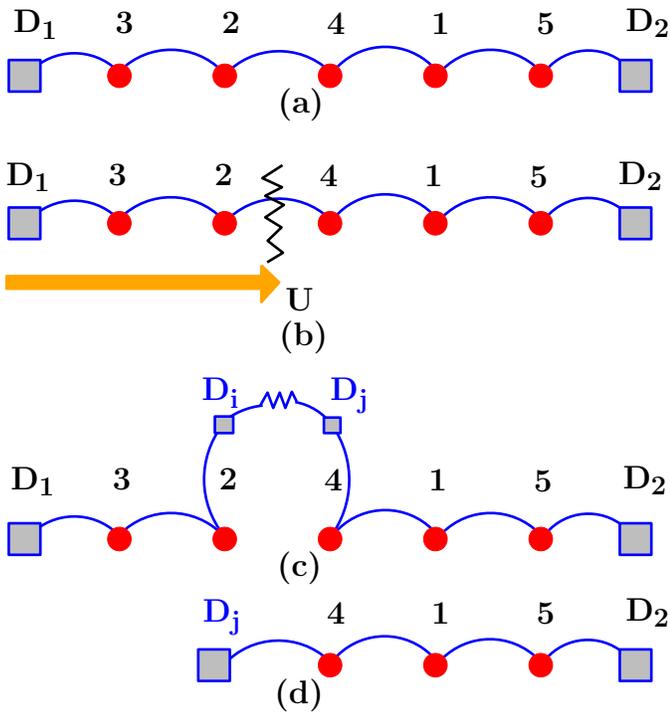}
	\caption{Tour repair algorithm for the TSP-based heuristic to generate feasible solutions to FCURP-MRS from TSP tours of the targets. (a) A subsequence of the tour with exactly two refueling sites, one at the start and other at the end. (b) Detection of fuel violation. (c) Compute indirect path to fix the fuel violation (d) Recursively call the repair algorithm on the unprocessed subsequence. }\label{fig:heuristic}	
\end{figure}

The initial TSP tour may be considered as a permutation of all targets with the first and last points on the tour being the starting depot. The input to Algorithm \ref{algo:heuristicRepair} is a sequence of targets and exactly two refueling sites, one at the beginning and the other at the end (Figure \ref{fig:heuristic}(a)). Algorithm \ref{algo:heuristicRepair} is a recursive routine that traverses the UAV route starting at the initial refueling site while keeping track of fuel consumption. It finds the first fuel capacity violation (lines \ref{line:whileFuel} - \ref{line:endwhileFuel}), if any, and calls the $indirectPath$ routine (Algorithm \ref{algo:indirectPath}) to fix the violation (see Figure \ref{fig:heuristic}(b)). The $indirectPath$ routine takes as input the two targets, $t_i$ and $t_j$ that constitute the violation, $s_{mrv}$, the most recently visited depot, $U_{rem}$, amount of fuel remaining at $t_i$ and $S$, the set of all refueling sites. It computes a feasible \emph{indirect path} comprising only of refueling sites between the two targets as shown in Figure \ref{fig:heuristic}(c). Once a feasible path is computed, target $t_i$ is frozen and is not processed again. Algorithm \ref{algo:heuristicRepair} then recursively calls itself (line \ref{line:recursiveCall}) on the sub sequence starting from the last depot on the feasible path with $t_j$ as the first target on the sequence (see Figure \ref{fig:heuristic}).

To compute the feasible path, Algorithm \ref{algo:indirectPath} builds a graph, $G_i$, with vertices as the union of the set of depots and the two targets. Edges in the graph exist between neighboring depots. $t_i$ is connected to depots that lie in the intersection set of reachable depots from $t_i$ and neighbors of the most recently visited depot. $t_j$ is connected to all its neighboring depots. It then computes the shortest path between $t_i$ and $t_j$ in $G_i$ and returns this as the feasible path to fix the violation. If the graph $G_i$ is not connected then Algorithm \ref{algo:indirectPath} cannot compute a feasible path. As mentioned in Section \ref{sec:siteSelection}, the set of refueling sites form a connected component and each target must be reachable from at least one depot. Hence if $G_i$ is disconnected, then $t_i$ is the disconnected vertex within $G_i$. In this case, Algorithm \ref{algo:heuristicRepair} backtracks to the previous target (line \ref{line:backTrack}). Now, $t_i$ becomes $t_j$ and the previous target becomes $t_i$. It keeps backtracking until it finds a pair of targets $t_i$ and $t_j$ such that it can compute an indirect path from $t_i$ to $t_j$. It may be noted that, except for the first iteration, the repairing algorithm would backtrack only up to the first target. It would always find a feasible path from the first target in the sequence to the second target. This is because the first target is atmost $U/2$ distance away from $s_{mrv}$ (Algorithm \ref{algo:indirectPath}, line \ref{indirectPath:tjEdges}). Hence in each iteration of the repairing algorithm, at least one target is added to the set of frozen targets. The heuristic will compute a feasible solution in at most $|T|$ iterations of the repairing algorithm and each iteration can make at most $|T|$ calls to the \emph{indirectPath} routine. Hence, the computation time of the heuristic is polynomial in the number of targets.

 \begin{algorithm}
 \caption{indirectPath}\label{algo:indirectPath}
 \begin{algorithmic}[1]
\State Build graph $G_i$ $\equiv (V,E)$\label{line:buildGraph}
\State $V \equiv S \cup \{t_i,t_j\}$
\State add edges between all neighboring depots \label{line:depotEdges}
\State add edges from $t_i$ to all reachable depots that are also neighbors of $s_{mrv}$ \label{line:mrvEdges}
\State add edges from $t_j$ to all depots within $U/2$ distance\label{indirectPath:tjEdges}
\State \Return $\mathcal{P}\equiv$ shortest path from $t_i$ to $t_j$ in $G_i$
 \end{algorithmic}
 \end{algorithm}
\section{Simulation Results}\label{sec:simResults}

\begin{figure*}[ht]
\begin{subfigure}{0.5\linewidth}
	\centering
	\includegraphics[scale = 0.3]{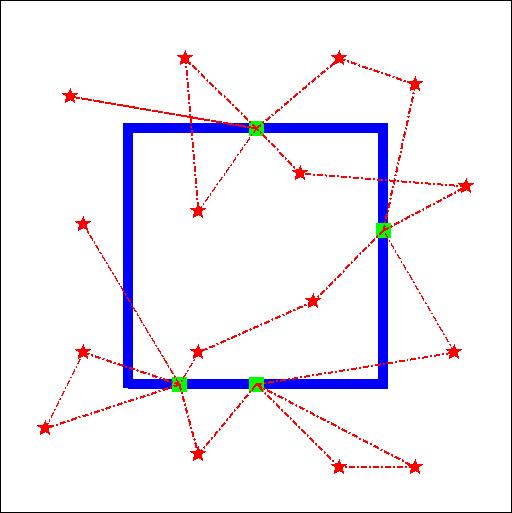}
	\caption{}\label{fig:simRoadNw1}	
\end{subfigure}
\begin{subfigure}{0.5\linewidth}
	\centering
	\includegraphics[scale = 0.3]{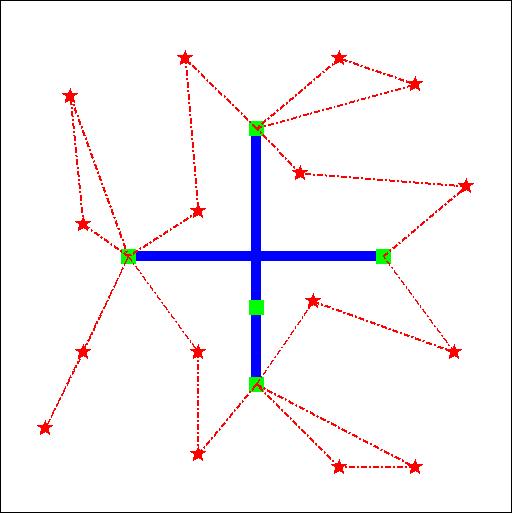}
	\caption{}\label{fig:simRoadNw2}	
\end{subfigure}
\caption{Sample instances used in the simulations. Blue lines represent the road network, red lines represent the UAV route, stars are the targets and green squares represent refueling sites. The two road networks used are shown in (a) and (b). The figures show sample tours for the UAV as generated by the solver.}\label{fig:simRoadNw}
\end{figure*}

In this section, the results of different solution approaches on simulated instances of FCURP-MRS are discussed. The section gives a comparison of solutions generated by the branch-and-cut algorithm for the two MILP formulations presented in Sec. \ref{sec:routePlanning}, TSP-based heuristic algorithm presented in Sec. \ref{sec:heuristic} and a baseline greedy algorithm \cite{ICUASself}.

\subsection{Simulation Setup}

Two different computing systems were used for the simulations. System 1 is an HP Spectre 360 laptop with Intel i7 7500U processor with 2 cores and 16 GB RAM running Windows 10. System 2 is a High Performance Cluster node with 20 cores and 96 GB RAM running Cent OS 6.5. The number of cores assigned to each process was dynamically decided by the HPC scheduler and was not fixed. Small computation activity was done on System 1 and simulations that needed large computation time and resources were executed on System 2.

The site selection algorithm (Algorithm \ref{algo:DCHS}) was implemented in Matlab and executed on System 1. MILP formulations were implemented in C++, using the traditional branch-and-cut framework and solver callback functionality of IBM ILOG CPLEX library version 12.7 and executed on System 2. A computational time limit of $7200$ seconds was imposed on every run of the branch-and-cut algorithm. The heuristic was also implemented in C++ and executed on System 1. The baseline greedy algorithm \cite{ICUASself} was coded in Matlab and executed on System 1. All implementations were coded in a serial flow. The solutions generated by the TSP-based heuristic were also input to the MILP solver as warm start for both formulations and the program was allowed to run for $7200$ seconds.

We report computation time and relative gap of the solution from the lower bound as computed by CPLEX. The performance of all algorithms was tested with randomly generated test instances as described in the following section.

\begin{table}
\centering
\begin{tabular}{lcccc}
\toprule
 \multirow{2}{*}{$|T|$} &\multicolumn{2}{c}{edge-based} & \multicolumn{2}{c}{node-based}\\
 &road-nw 1  &road-nw 2  &road-nw 1  &road-nw 2 \\
\midrule
$9$ & $100 $  & $86.67 (13.33)$   & $100$      & $86.67 (13.33)$\\
$16$ & $95.83$ & $79.17 (20)$  &  $99.16$  & $80 (20)$\\
$25$ & $13.33$       & $0 (33.33)$     & $35$       & $0 (33.33)$\\
$36$ & $0$       & $0 (33.33)$   & $0$     & $0 (33.33)$\\
\bottomrule
\end{tabular}
\caption{Percentage of instances solved to optimality by the MILP solver within $7200$ seconds. Numbers in braces represent the percentage of infeasible instances, if any.}
\label{tab:1}
\end{table}

\subsection{Instance Generation} \label{subsec:instances}

For all simulations, a square environment of size $20 \times 20$ square kilometers (km) is used. The round trip distance between the two farthest points in the environment is approximately $57$ km; this distance is greater than the maximum distance a small UAV can travel in one flight time \cite{Blyenburgh2007}. The value of $U$ is varied in the range $15$ to $25$ km and $R$ is varied in the range $10$ to $15$ km, both in steps of $5$. Two different road networks, as shown in Figure \ref{fig:simRoadNw}, are used. To achieve area coverage, an $n\times n$ grid with $n$ ranging from $3$ to $10$, where $n$ denotes the grid size, is placed in the environment. Targets are placed at the grid centers and the number of target is varied in the range $9$ to $100$. $20$ instances are created for each configuration by selecting target locations with uniform distribution.
 
The two different road networks used, are shown in Figure \ref{fig:simRoadNw}. Road Network 1 is very expansive within the environment and the mean shortest distance of any point in the environment from the road network is small. While in the case of Road Network 2, the roads are much scarce and the mean shortest distance is much higher, specifically the points in each of the four corners of the environment are far off from the road network. The choice of particular road networks was made to include two representative extremes of road network coverage. In each case, the road network is discretized at a resolution of $1$ km to generate a set of candidate refueling sites. Then, the greedy algorithm presented in Section \ref{sec:siteSelection} is used to obtain a reduced set of refueling sites. The set of refueling sites along with target locations are given as input to the second stage to compute routes for UAV and RV.
\begin{table}
\centering
\begin{tabular}{lcccc}
\toprule
 \multirow{2}{*}{$|T|$} &\multicolumn{2}{c}{edge-based} & \multicolumn{2}{c}{node-based}\\
 &road-nw 1  &road-nw 2  &road-nw 1  &road-nw 2 \\
\midrule
9 & 100  & 86.67 (13.33)   & 100  & 86.67 (13.33)\\
16 & 100  & 80 (20)           & 100  & 80 (20)\\
25 & 100  & 66.67 (33.33)   & 100  & 66.67 (33.33)\\
36 & 100  & 66.67 (33.33)   & 100  & 66.67 (33.33)\\
\bottomrule
\end{tabular}
\caption{Percentage of instances for which the MILP solver could compute a feasible solution when given a warm start using the solutions generated by the TSP-based Heuristic. Numbers in braces represent the percentage of infeasible instances, if any. Feasible solutions were computed for all feasible instances by the heuristic.}
\label{tab:2}
\end{table}

\begin{table}
\centering
\begin{tabular}{lcccc}
\toprule
 \multirow{2}{*}{$|T|$} &\multicolumn{2}{c}{edge-based} & \multicolumn{2}{c}{node-based}\\
 &road-nw 1  &road-nw 2  &road-nw 1  &road-nw 2 \\
\midrule
9 & 0.94    & 0.90     &   0.83    &  0.61\\
16 & 1.00    & 1.00     &   0.98    &  0.99\\
25 & 10.85  & 21.84    & 10.10    & 21.18\\
36 & 18.06  & 28.84    & 21.69    & 34.34\\
\bottomrule
\end{tabular}
\caption{Relative gap percentage from lower bound as computed by CPLEX (computed only for feasible instances) for feasible solutions computed by using the TSP-based heuristic solution as warm start to the formulations.}
\label{tab:3}
\end{table}

\begin{figure*}
\centering
\begin{subfigure}{0.32\linewidth}
	\centering
	\includegraphics[width =  1.125\linewidth]{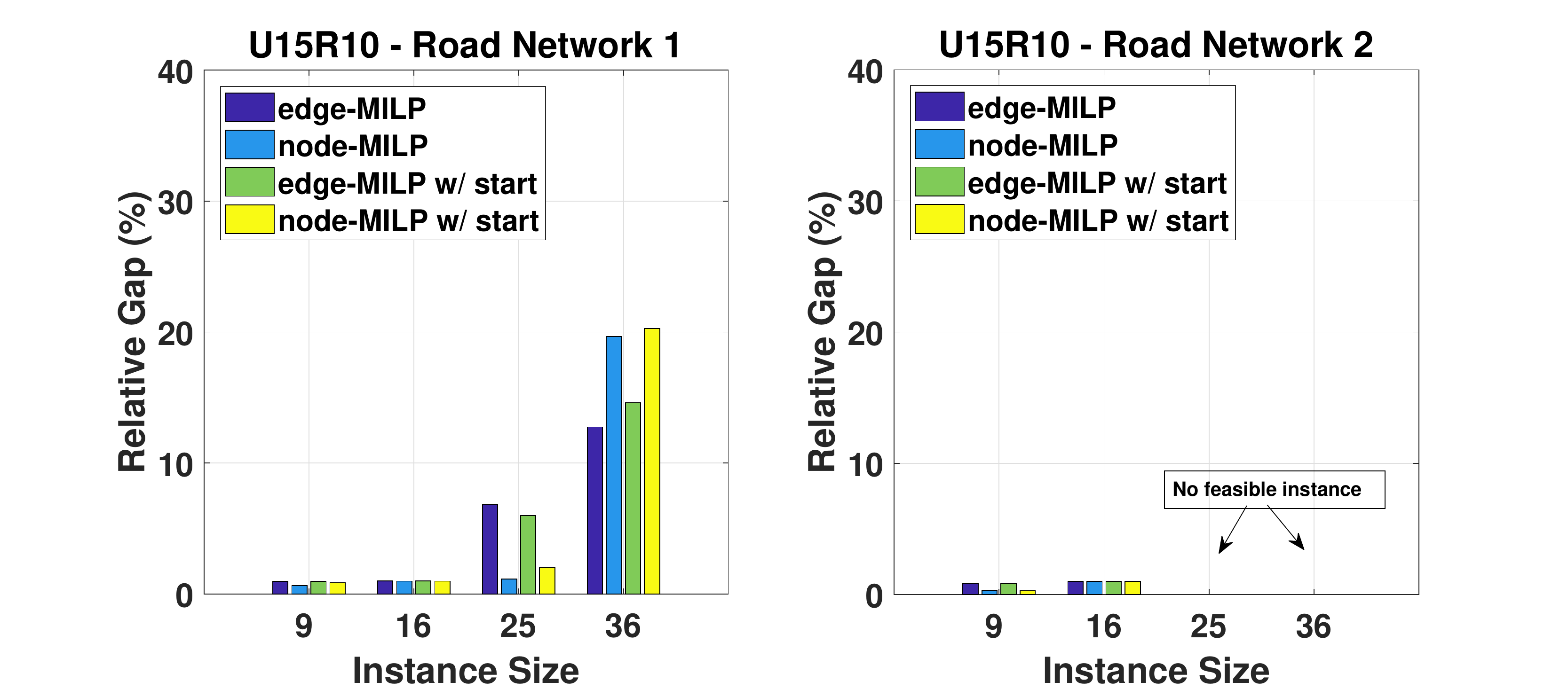}
	\caption{}\label{fig:simRoadNw1}	
\end{subfigure}
\begin{subfigure}{0.32\linewidth}
	\centering
	\includegraphics[width =  1.125\linewidth]{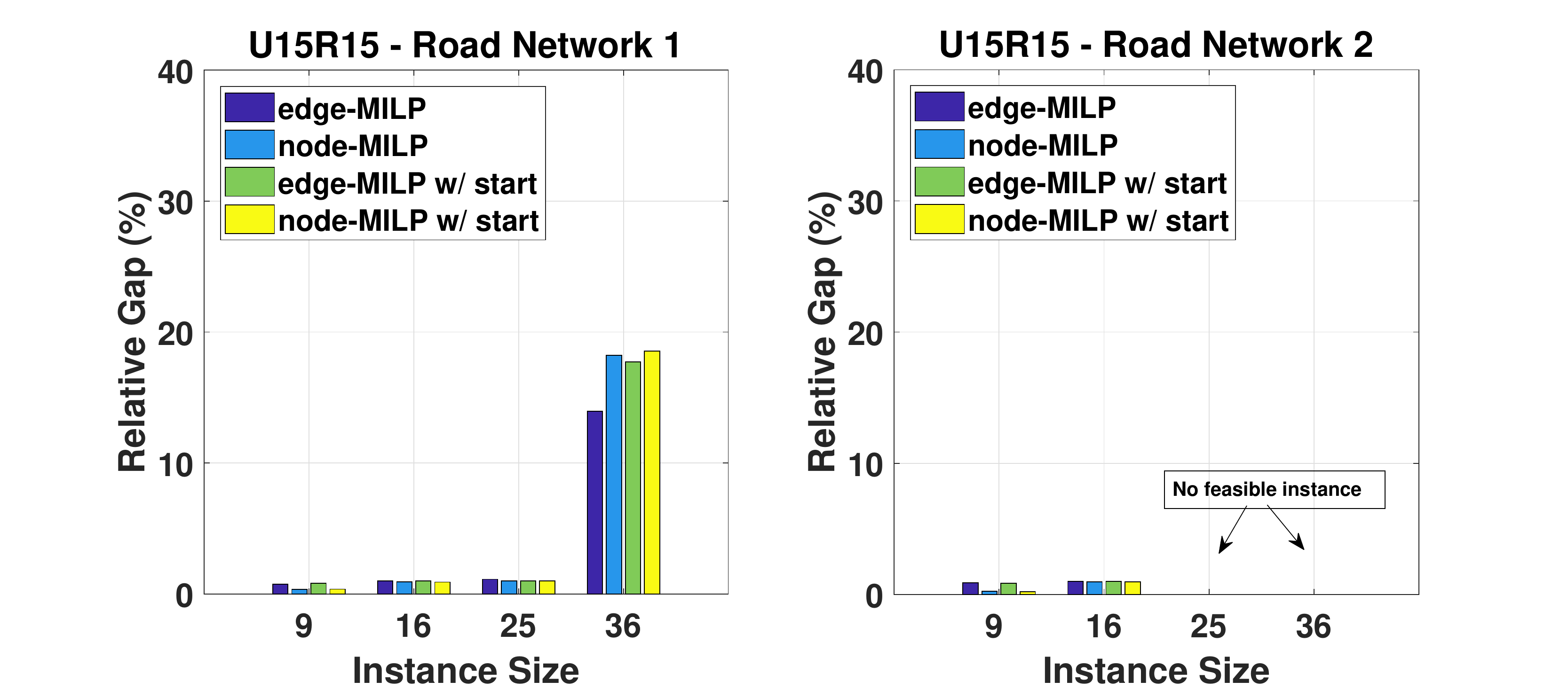}
	\caption{}\label{fig:simRoadNw2}	
\end{subfigure}
\begin{subfigure}{0.32\linewidth}
	\centering
	\includegraphics[width =  1.125\linewidth]{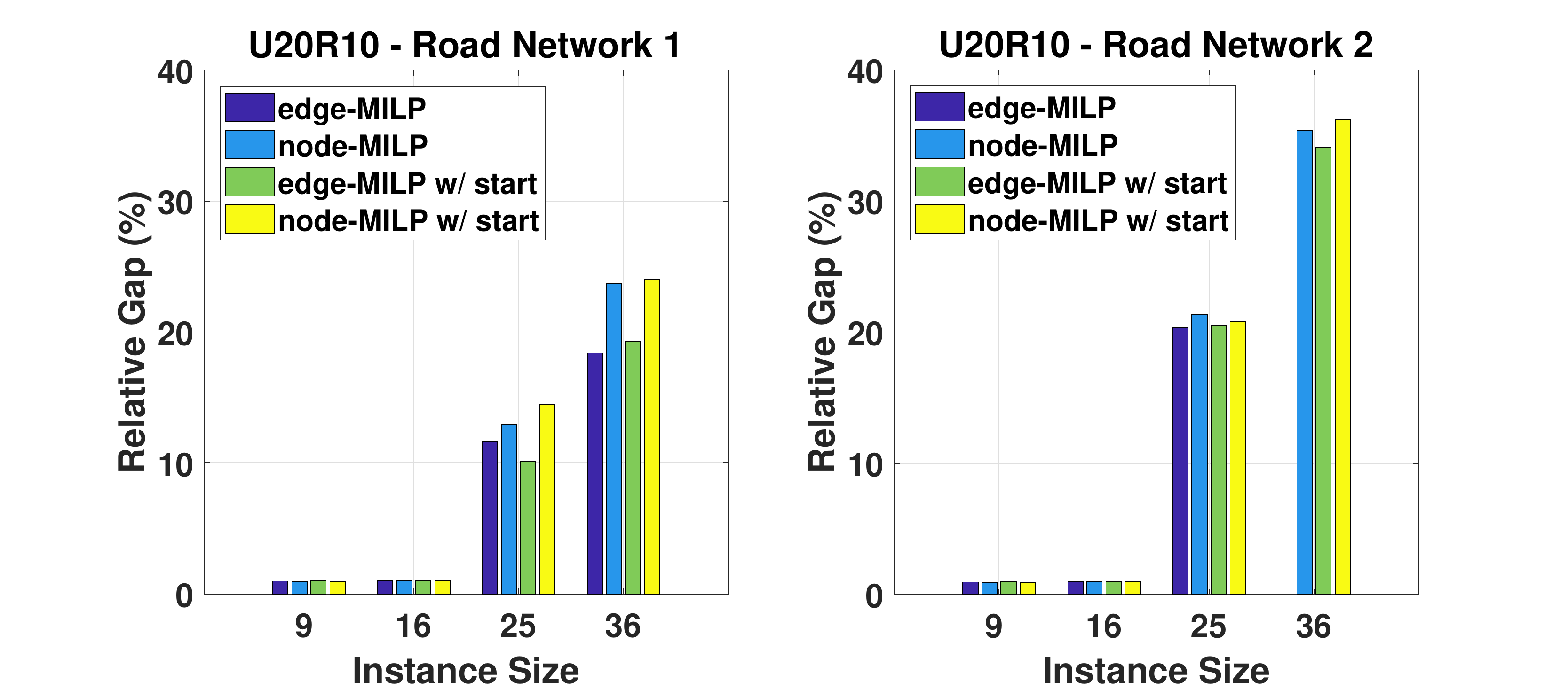}
	\caption{}\label{fig:simRoadNw2}	
\end{subfigure}
\begin{subfigure}{0.32\linewidth}
	\centering
	\includegraphics[width =  1.125\linewidth]{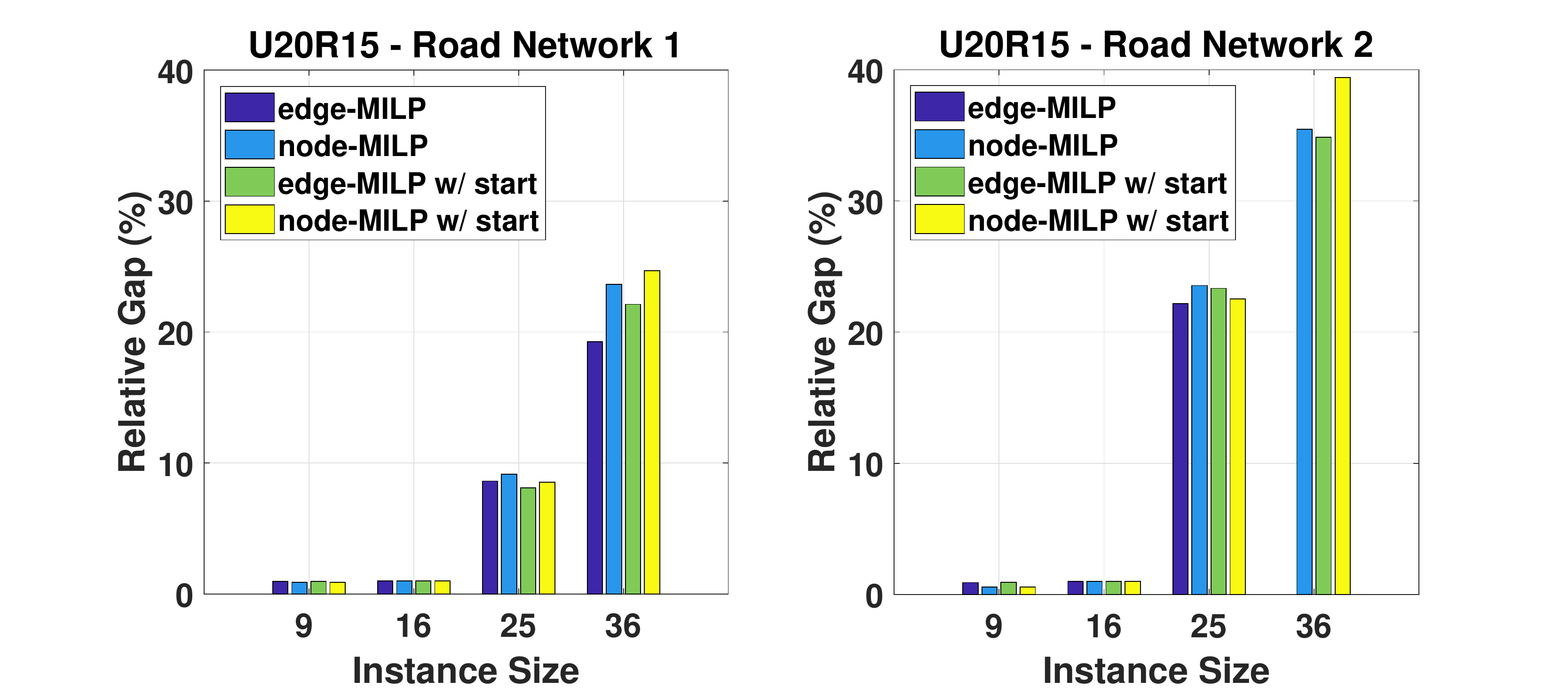}
	\caption{}\label{fig:simRoadNw1}	
\end{subfigure}
\begin{subfigure}{0.32\linewidth}
	\centering
	\includegraphics[width =  1.125\linewidth]{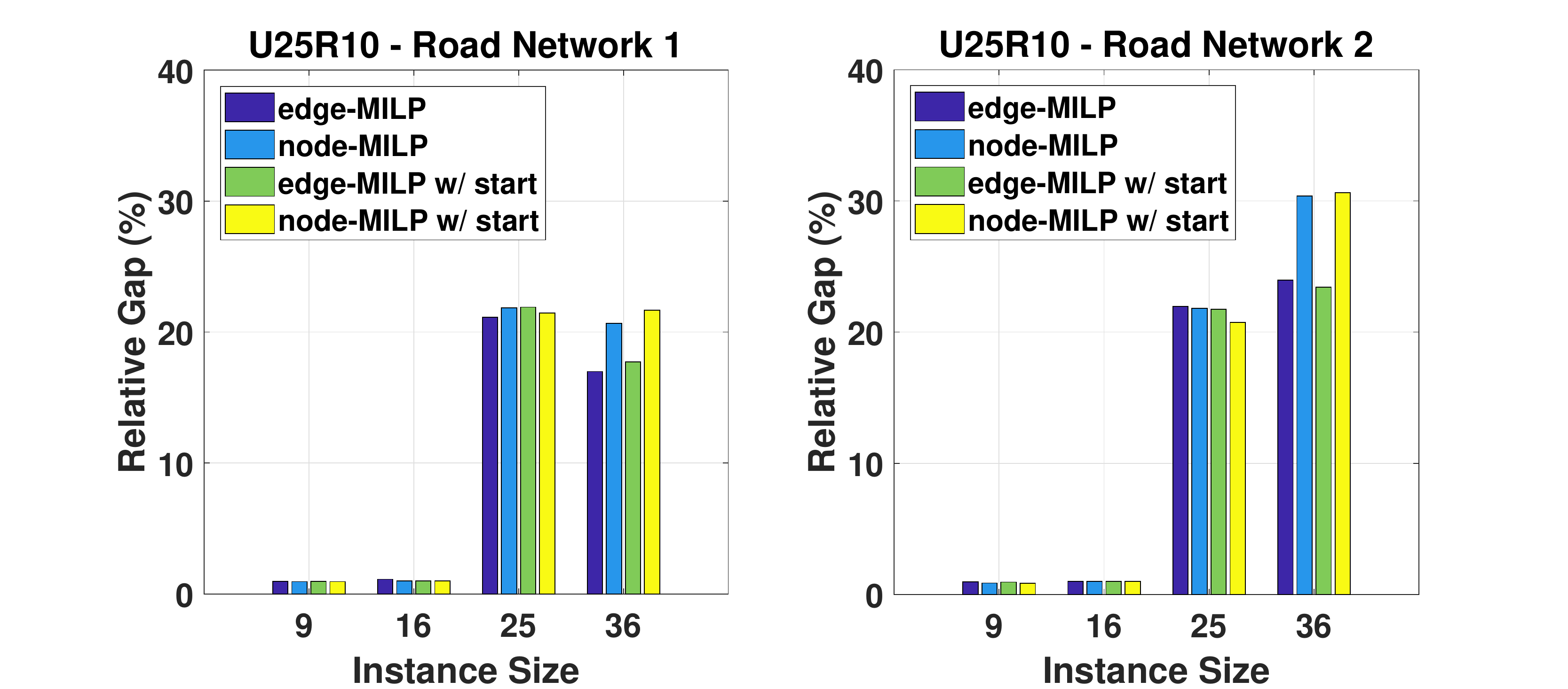}
	\caption{}\label{fig:simRoadNw2}	
\end{subfigure}
\begin{subfigure}{0.32\linewidth}
	\centering
	\includegraphics[width =  1.125\linewidth]{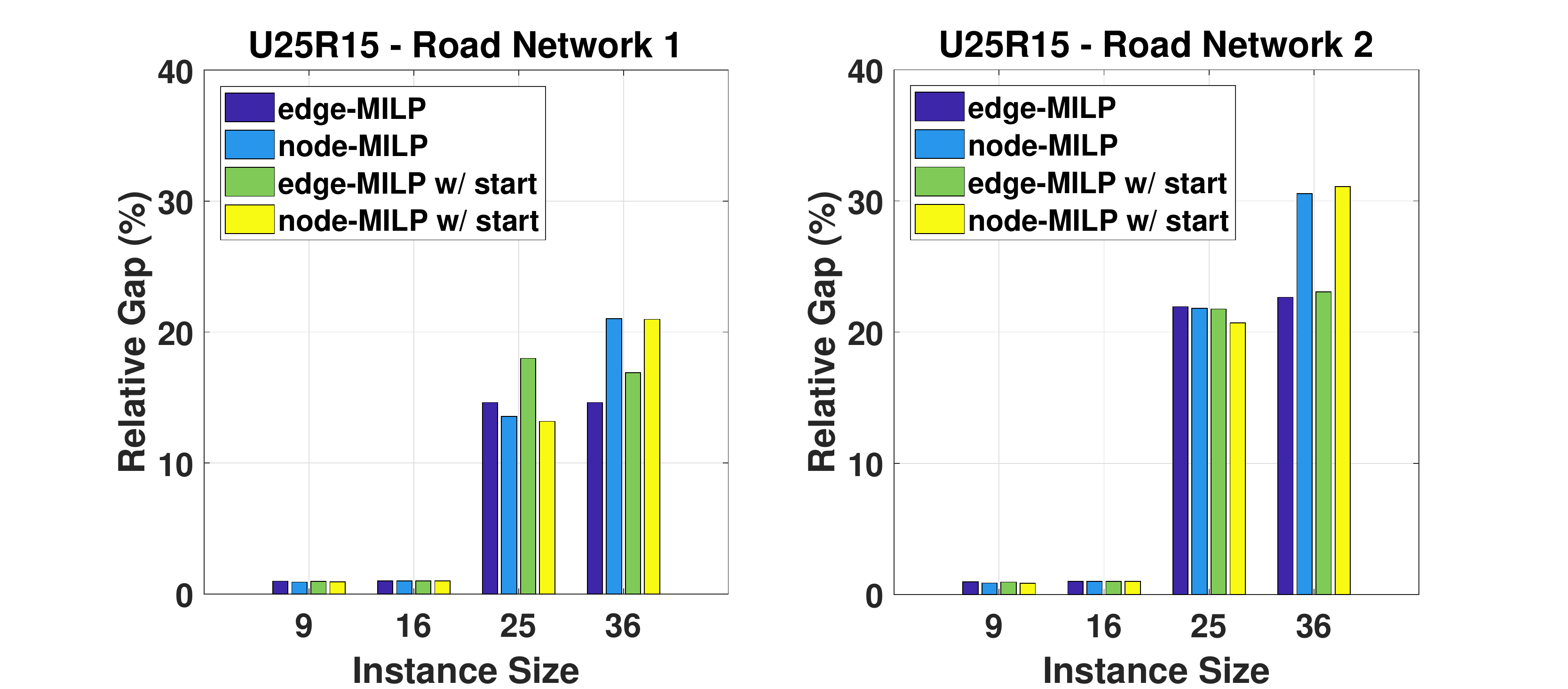}
	\caption{}\label{fig:simRoadNw2}	
\end{subfigure}
\caption{Relative Gap of the MILP solution from the lower bounds as generated by the MILP solver.}\label{fig:gapMILP}
\end{figure*}

\begin{figure*}
\centering
\begin{subfigure}{0.32\linewidth}
	\centering
	\includegraphics[width = 1.125\linewidth]{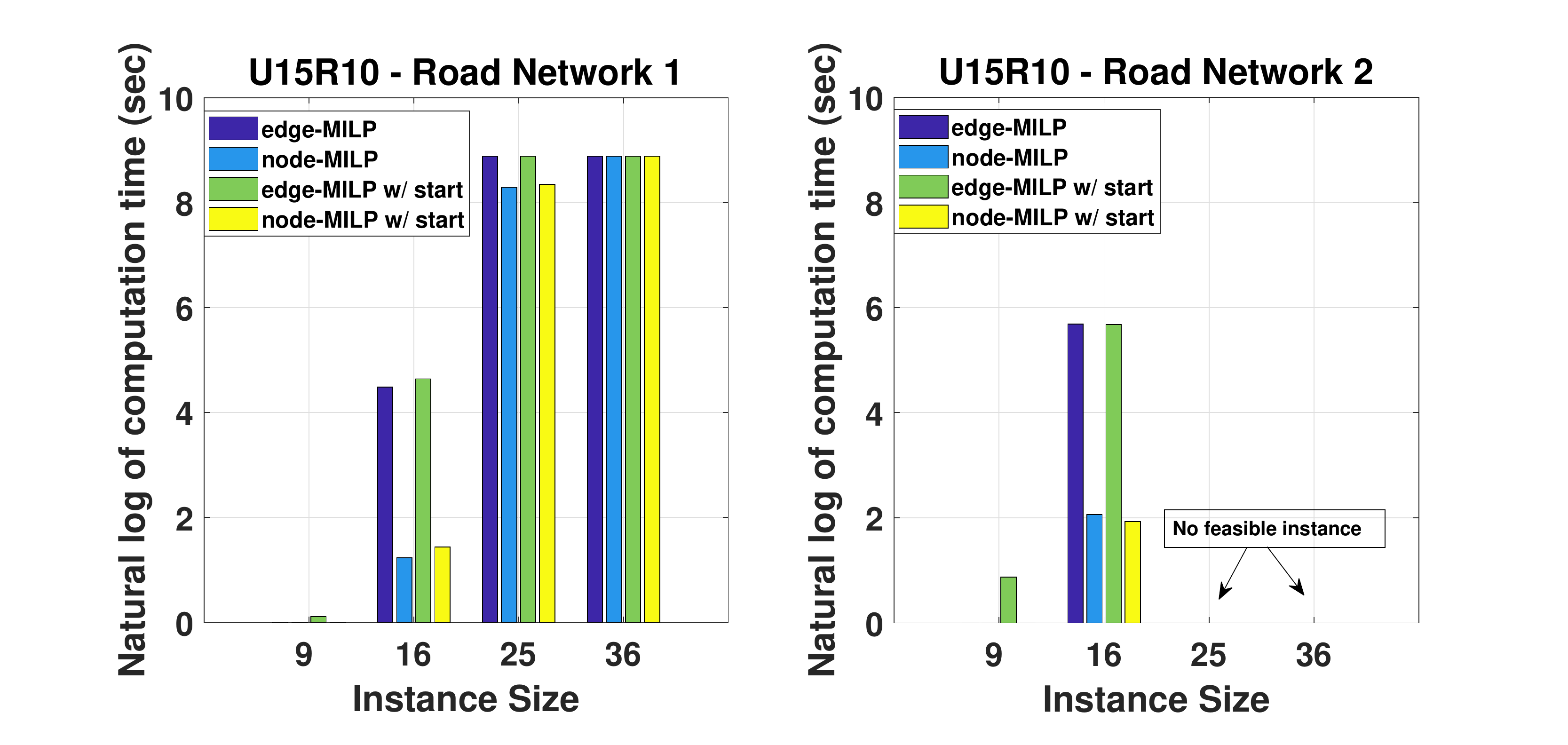}
	\caption{}\label{fig:simRoadNw1}	
\end{subfigure}
\begin{subfigure}{0.32\linewidth}
	\centering
	\includegraphics[width = 1.125\linewidth]{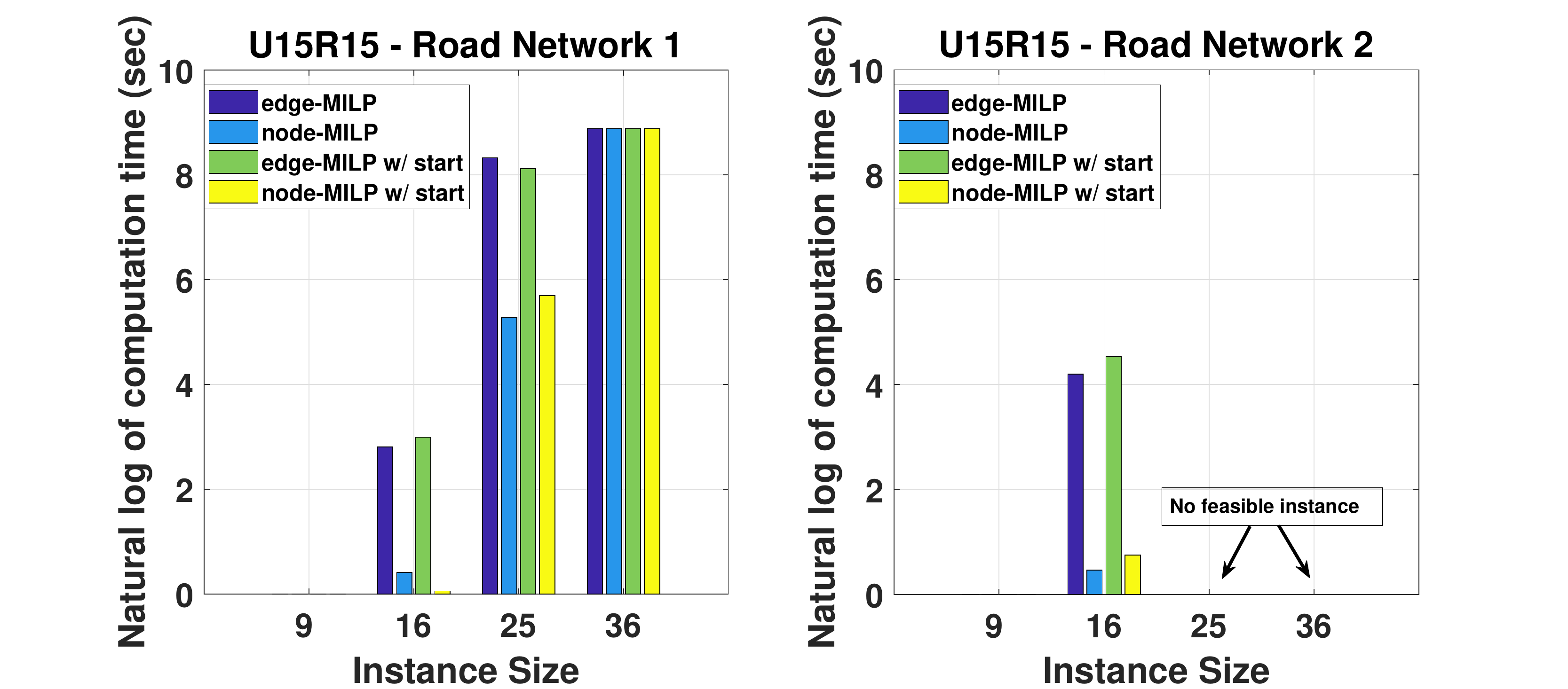}
	\caption{}\label{fig:simRoadNw2}	
\end{subfigure}
\begin{subfigure}{0.32\linewidth}
	\centering
	\includegraphics[width = 1.125\linewidth]{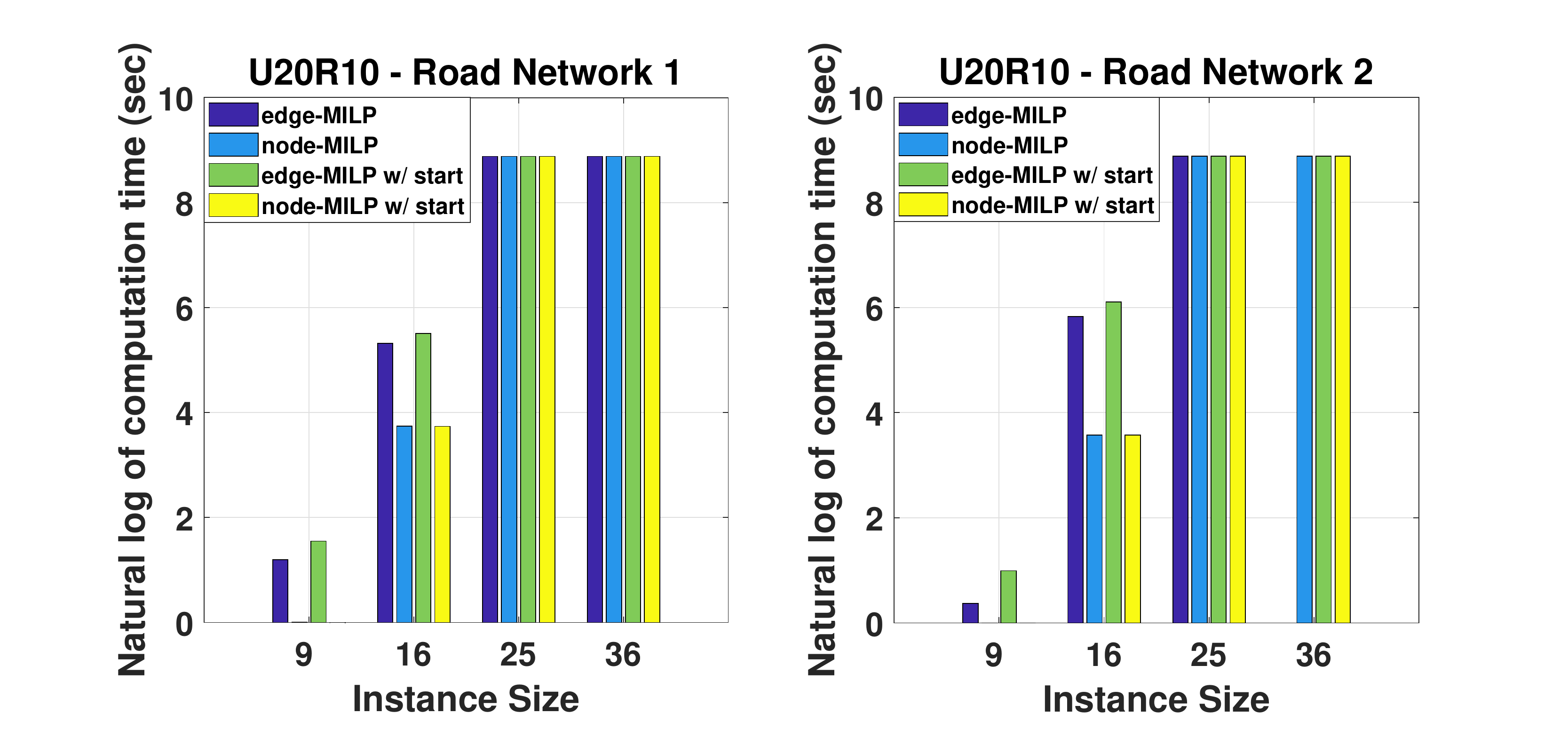}
	\caption{}\label{fig:simRoadNw2}	
\end{subfigure}
\begin{subfigure}{0.32\linewidth}
	\centering
	\includegraphics[width = 1.125\linewidth]{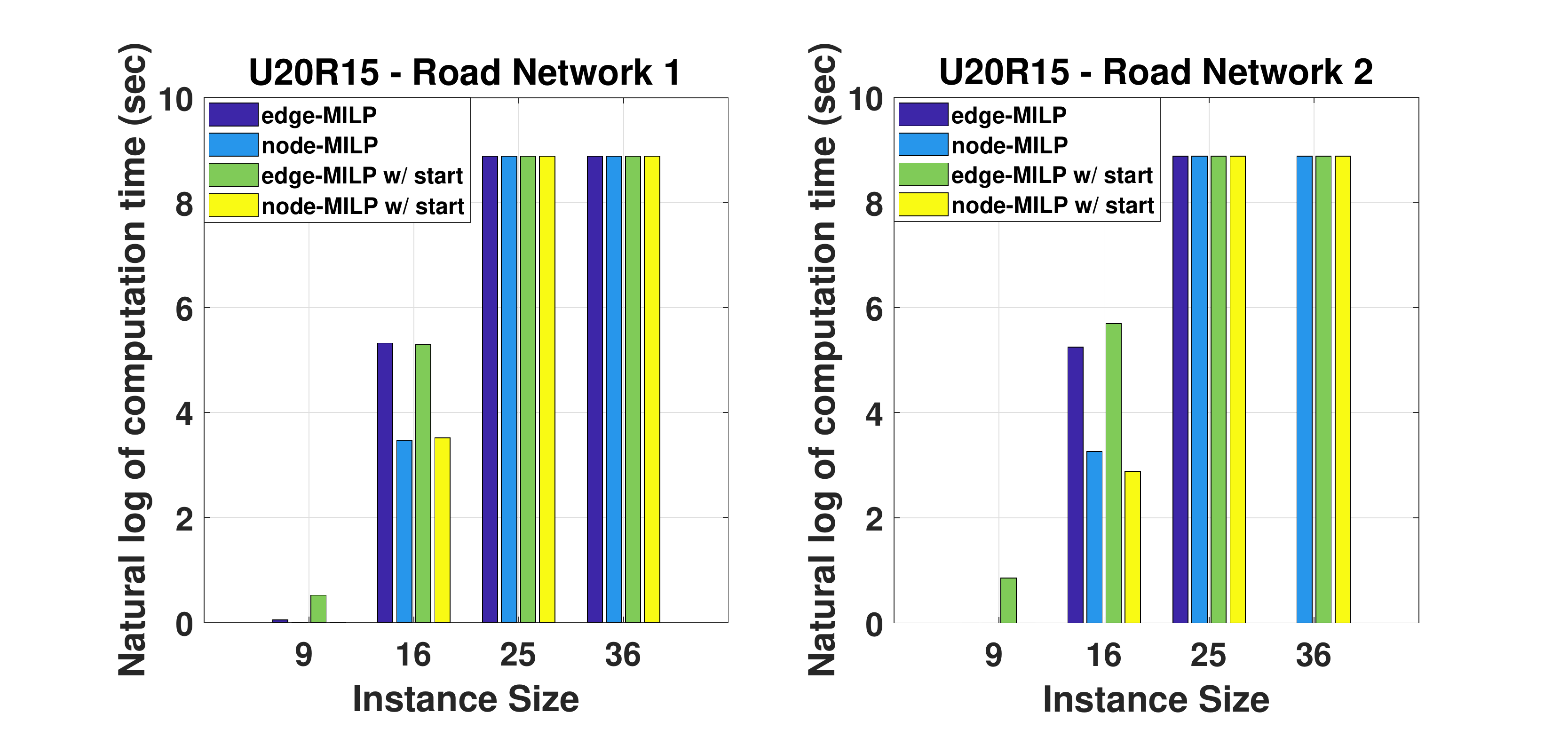}
	\caption{}\label{fig:simRoadNw1}	
\end{subfigure}
\begin{subfigure}{0.32\linewidth}
	\centering
	\includegraphics[width = 1.125\linewidth]{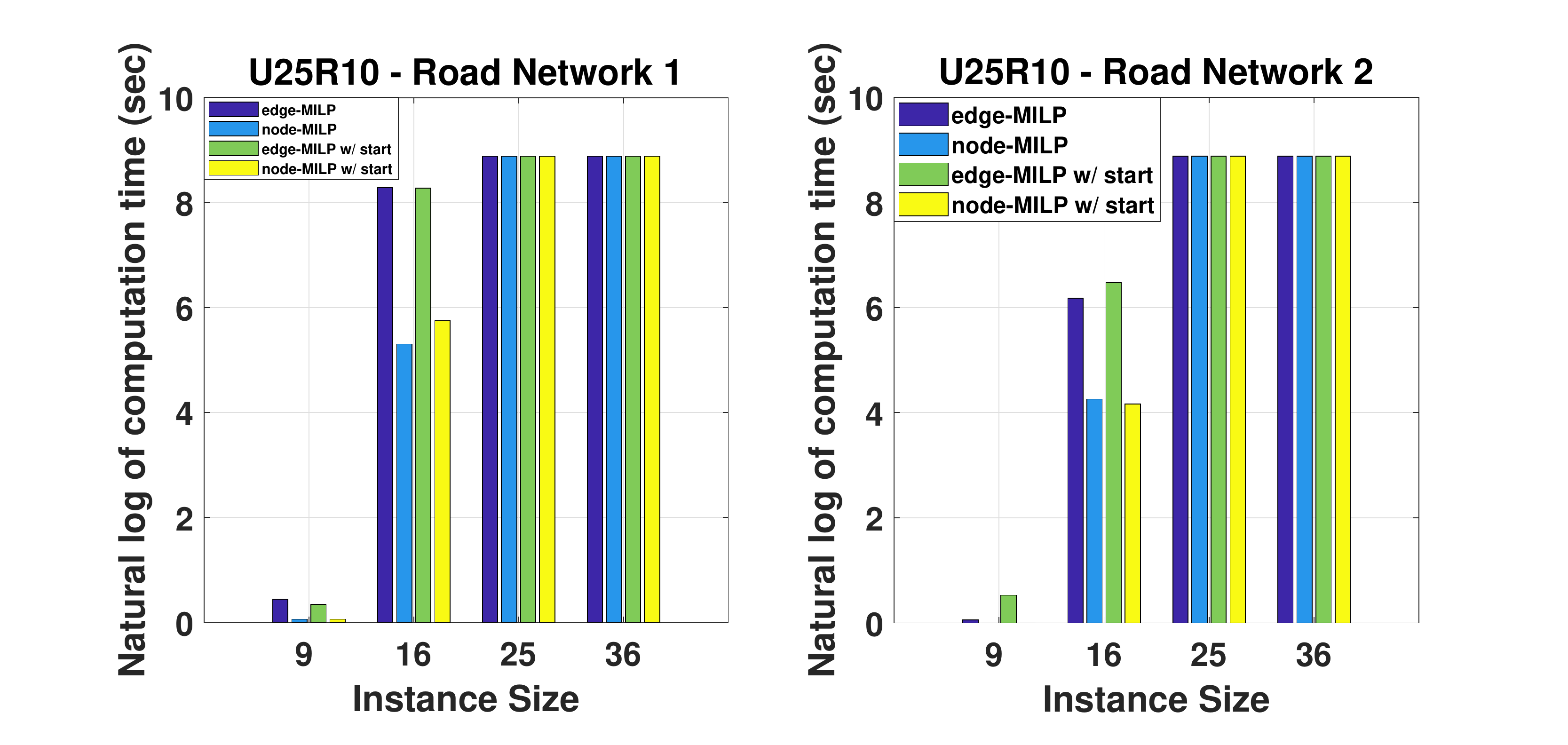}
	\caption{}\label{fig:simRoadNw2}	
\end{subfigure}
\begin{subfigure}{0.32\linewidth}
	\centering
	\includegraphics[width = 1.125\linewidth]{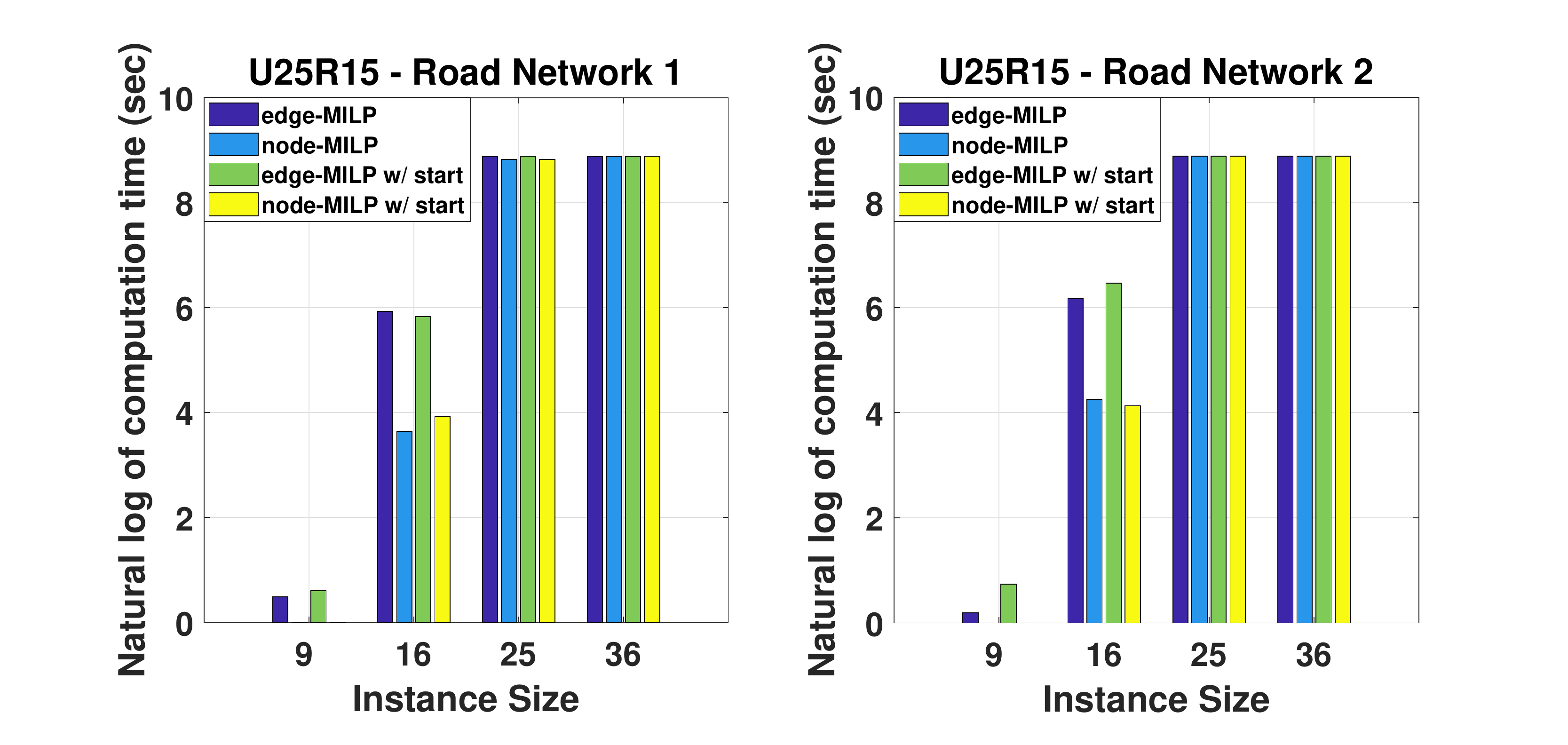}
	\caption{}\label{fig:simRoadNw2}	
\end{subfigure}
\caption{Computation time of the MILP solver for the simulation instances using edge and node based formulations, with and without warm start. }\label{fig:timeMILP}
\end{figure*}

\subsection{Results}

Simulation instances generated as described in Section \ref{subsec:instances} were solved using six strategies as given: 
\begin{itemize}
\item Edge-based MILP
\item Node-based MILP
\item Edge-based MILP with warm start
\item Node based MILP with warm start
\item TSP-based Heuristic
\item Baseline Greedy Algorithm
\end{itemize}

The results of using each of these strategies and their comparisons are presented next. The MILP solvers were used to generate optimal solutions for benchmarking purposes. In the case, when the optimal solution was not found within stipulated time limit, lower bounds generated by the solver were used. The lower bounds give slightly magnified bounds, but help develop an understanding of the performance of an algorithm. Table \ref{tab:1} shows the percentage of instances that were solved to optimality for different values of $n$. The solver was not able to compute optimal solutions for any instances with more than $25$ targets ($n=5$), hence the table only shows the numbers for up to 36 targets. The fraction of instances that were solved to \emph{optimality} in the stipulated time period decreases drastically with increasing values of $n$. The numbers were slightly better for the node-based formulation. However, both formulations show a clear trend and do not scale well for larger instances. In case of road-network 2, some of the instances were infeasible due to the limited road network. The percentage of such instances is indicated in braces in the table. The site selection algorithm in stage one of the strategy was not able to compute a a feasible set of refueling sites for any of these instance.

\begin{figure*}
\centering
\begin{subfigure}{0.32\linewidth}
	\centering
	\includegraphics[width = 1.125\linewidth]{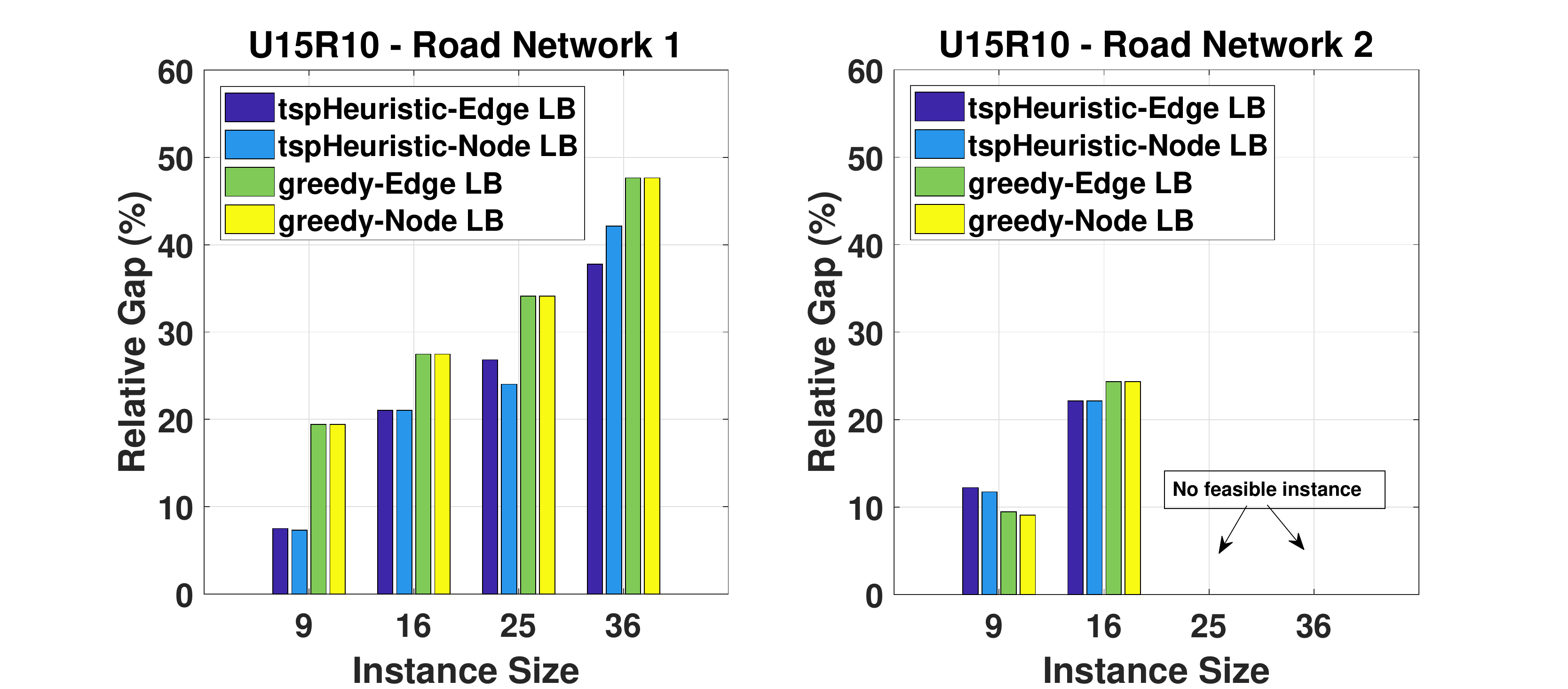}
	\caption{}\label{fig:simRoadNw1}	
\end{subfigure}
\begin{subfigure}{0.32\linewidth}
	\centering
	\includegraphics[width =  1.125\linewidth]{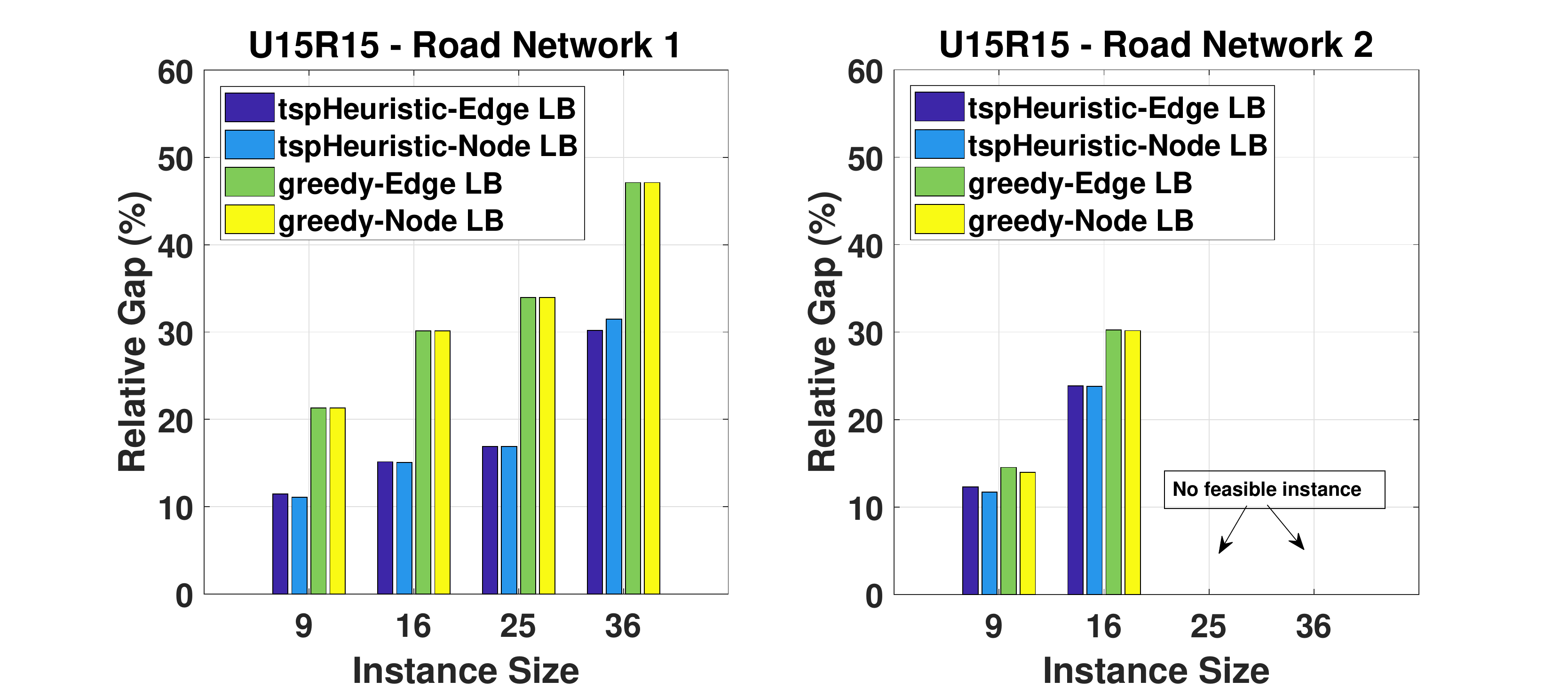}
	\caption{}\label{fig:simRoadNw2}	
\end{subfigure}
\begin{subfigure}{0.32\linewidth}
	\centering
	\includegraphics[width =  1.125\linewidth]{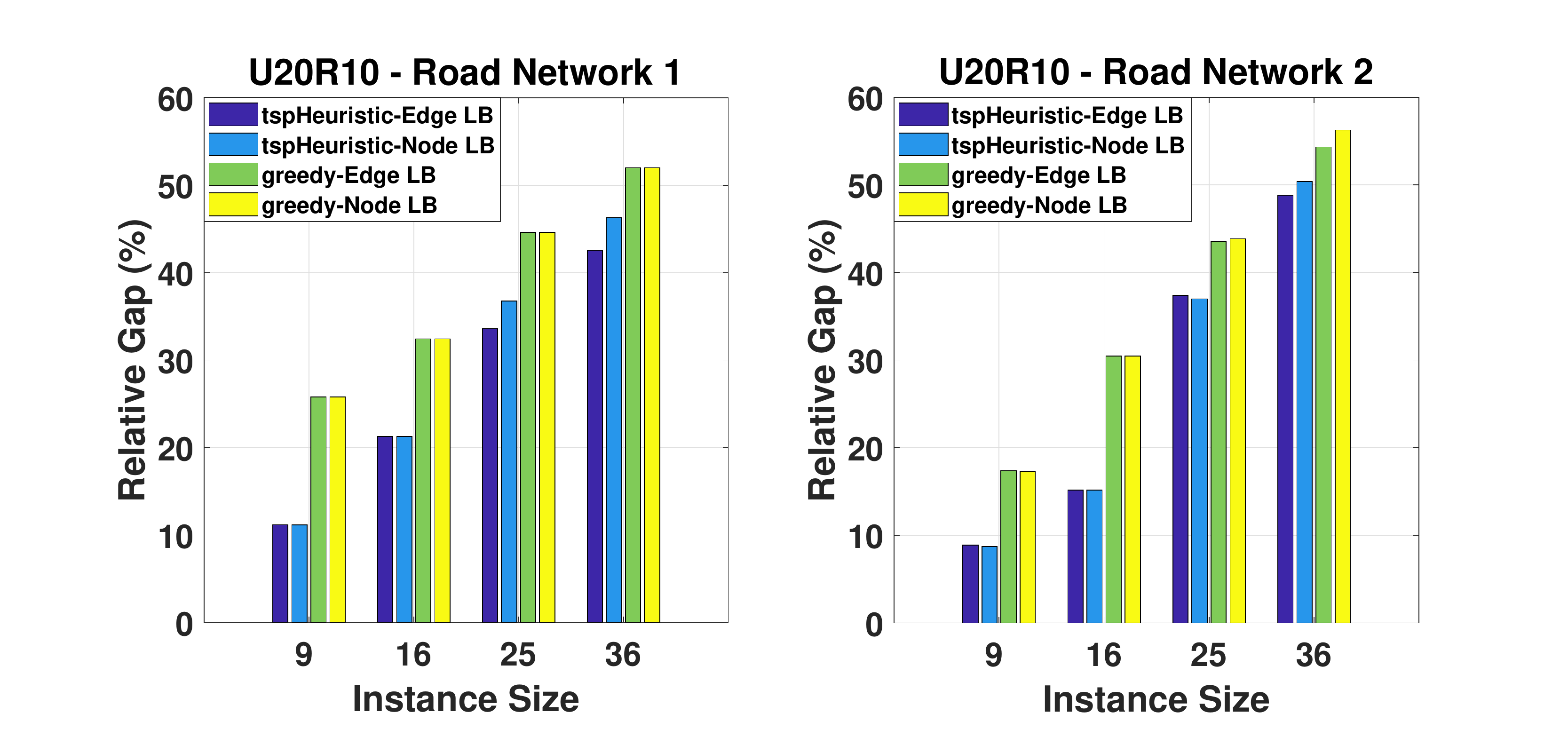}
	\caption{}\label{fig:simRoadNw2}	
\end{subfigure}
\begin{subfigure}{0.32\linewidth}
	\centering
	\includegraphics[width =  1.125\linewidth]{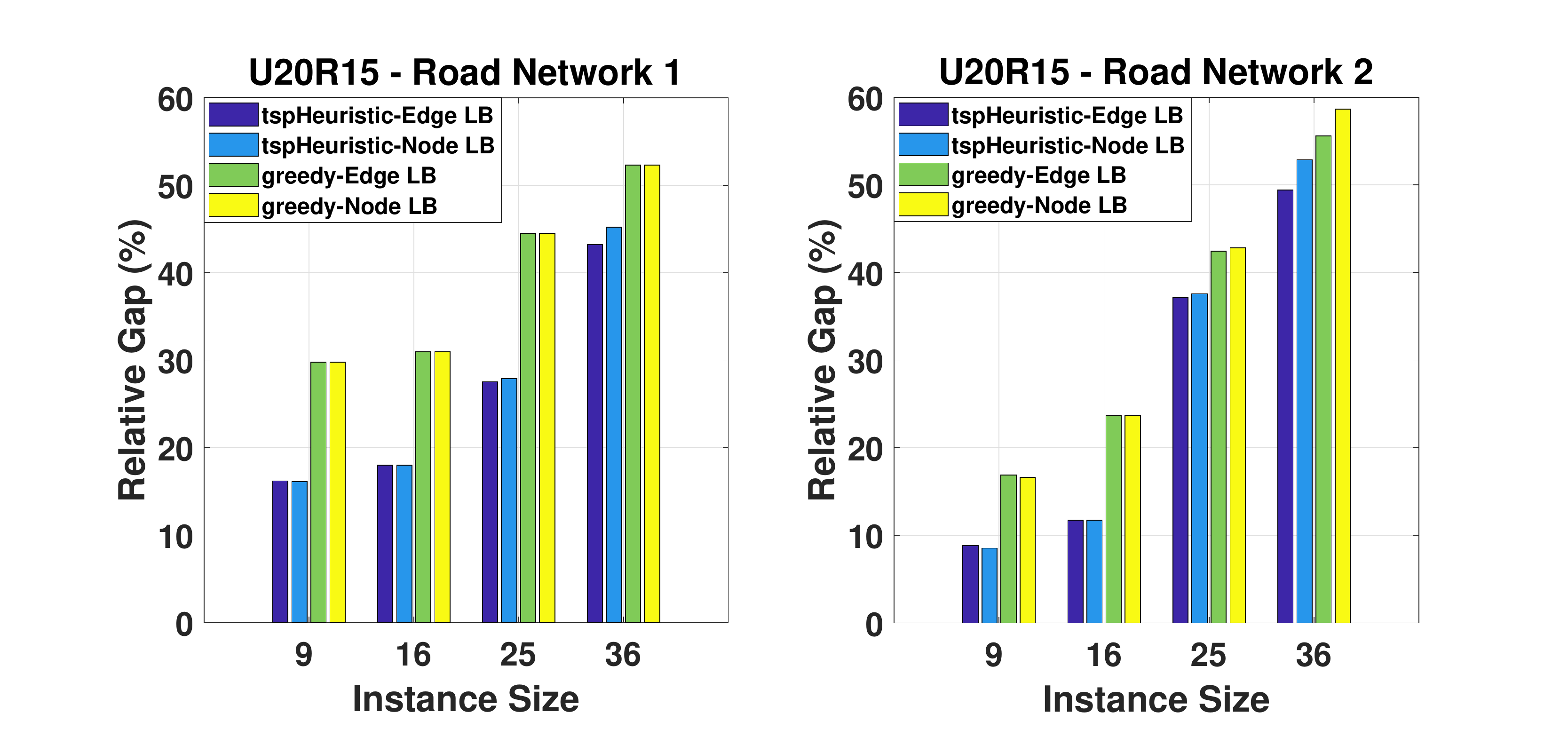}
	\caption{}\label{fig:simRoadNw1}	
\end{subfigure}
\begin{subfigure}{0.32\linewidth}
	\centering
	\includegraphics[width =  1.125\linewidth]{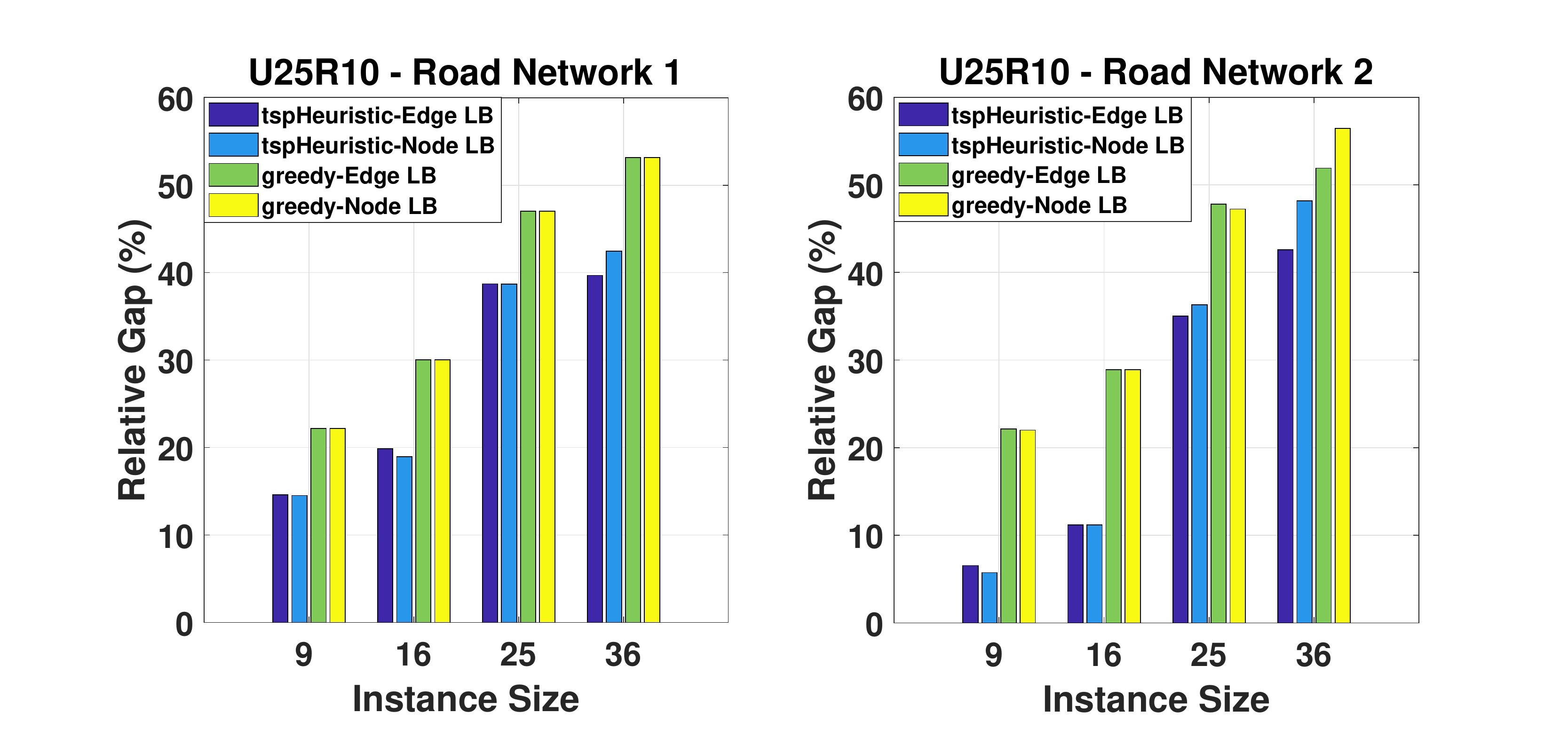}
	\caption{}\label{fig:simRoadNw2}	
\end{subfigure}
\begin{subfigure}{0.32\linewidth}
	\centering
	\includegraphics[width =  1.125\linewidth]{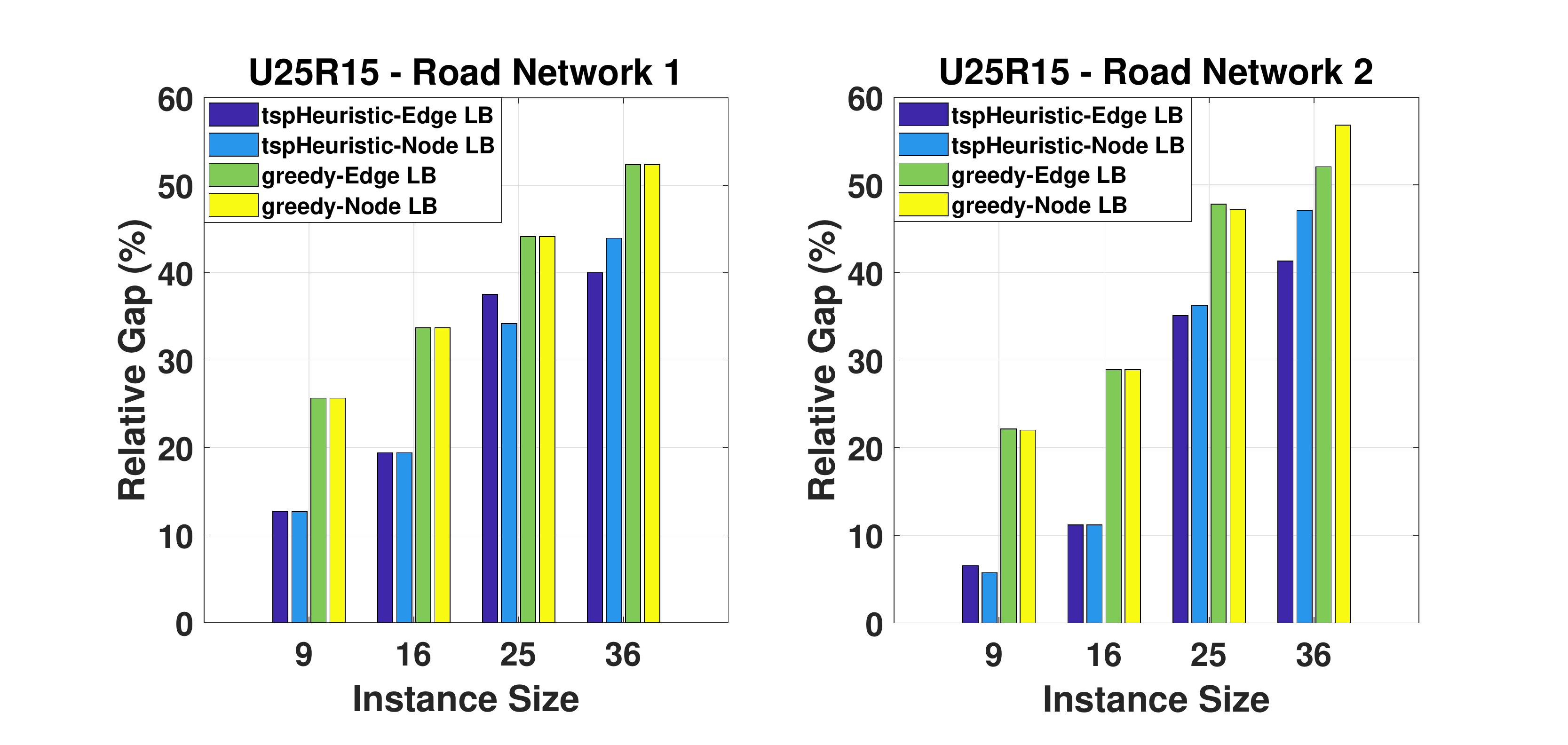}
	\caption{}\label{fig:simRoadNw2}	
\end{subfigure}
\caption{Solution quality in terms of relative gap of the heuristic solutions using lower bounds generated by the MILP solver for the edge and node based formulations.}\label{fig:gapHeuristic}
\end{figure*}

Tables \ref{tab:2} and \ref{tab:3} give the computation results for the MILP formulations when the solver was given a warm start with the solutions generated by the TSP-based heuristic. Table \ref{tab:2} reports the percentage of instances solved while Table \ref{tab:3} tabulates the relative gap of the solutions. This gap is computed as the percentage difference between the feasible solution and the lower bound as computed by the MILP solver. The solver was set to stop the optimization process at $1 \%$ relative gap. It may be observed, that while the solver was able to compute feasible solutions for all instances, the solution quality degrades rapidly with increase in instance size. It may also be observed that the gap is higher for road-network 2. This is attributed to the large value of the mean shortest distance of targets from the road network. This is also observed from the infographic in Figure \ref{fig:gapMILP}. The figure also gives insights on the performance of the two different MILP paradigms, namely edge-based and node-based. The edge-based formulations consistently has lower relative gap than the node-based formulations over all combinations of simulation parameters, namely $U$, $R$, $n$ and the road-network. At the same time, the node-based formulations is computationally more efficient, for the instances where it is able to compute the optimal, as seen in Figure \ref{fig:timeMILP}. This is because of the lower dimensionality of the node-based formulations. A counter intuitive insight from these figures is that the warm start does not improve the solution quality nor does it help the formulations to converge faster. This is not consistent with results for similar approaches for other combinatorial problems. The reason for these results are manifold. One, the heuristic solutions have significant gap (Figure \ref{fig:gapHeuristic}) and do not give much improved starting points for the solver. Two, this problem has a very large number of constraints and has a strong coupling between the routes for the two vehicles. For these reasons, the heuristic solutions do not improve the MILP formulation.

\begin{figure*}
\centering
\begin{subfigure}{0.32\linewidth}
	\centering
	\includegraphics[clip,trim=20mm 5mm 0mm 5mm, width =  1.125\linewidth]{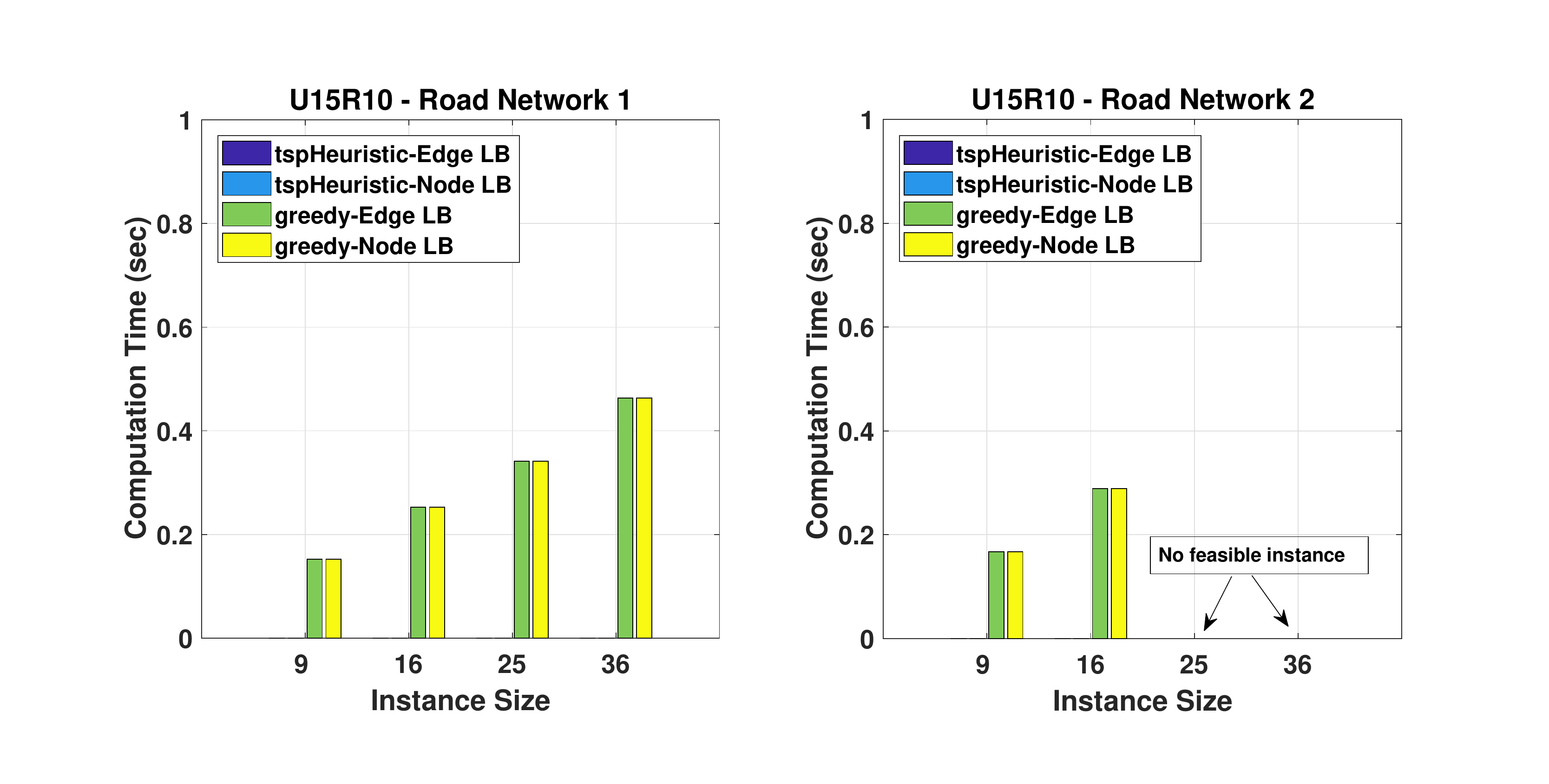}
	\caption{}\label{fig:simRoadNw1}	
\end{subfigure}
\begin{subfigure}{0.32\linewidth}
	\centering
	\includegraphics[clip,trim=20mm 5mm 0mm 5mm,width =  1.125\linewidth]{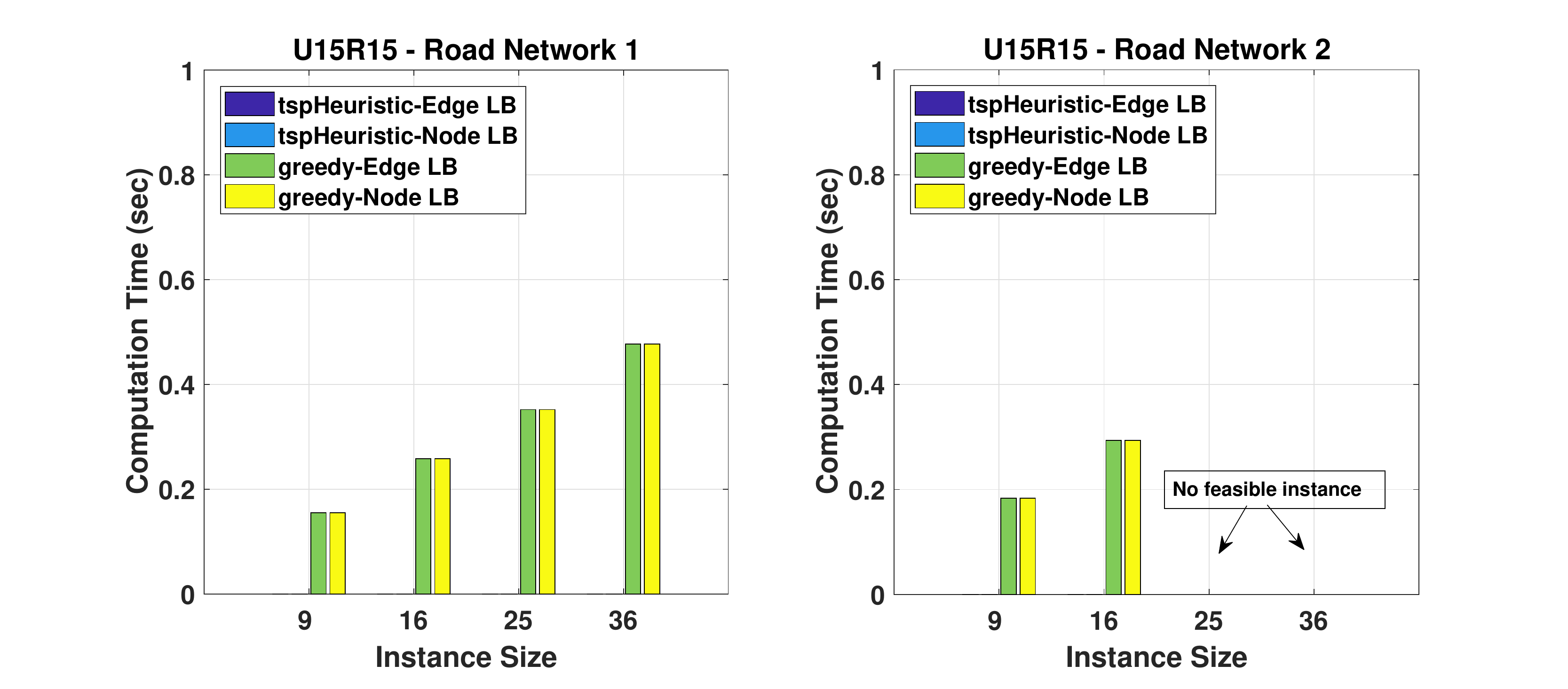}
	\caption{}\label{fig:simRoadNw2}	
\end{subfigure}
\begin{subfigure}{0.32\linewidth}
	\centering
	\includegraphics[clip,trim=20mm 5mm 0mm 5mm,width =  1.125\linewidth]{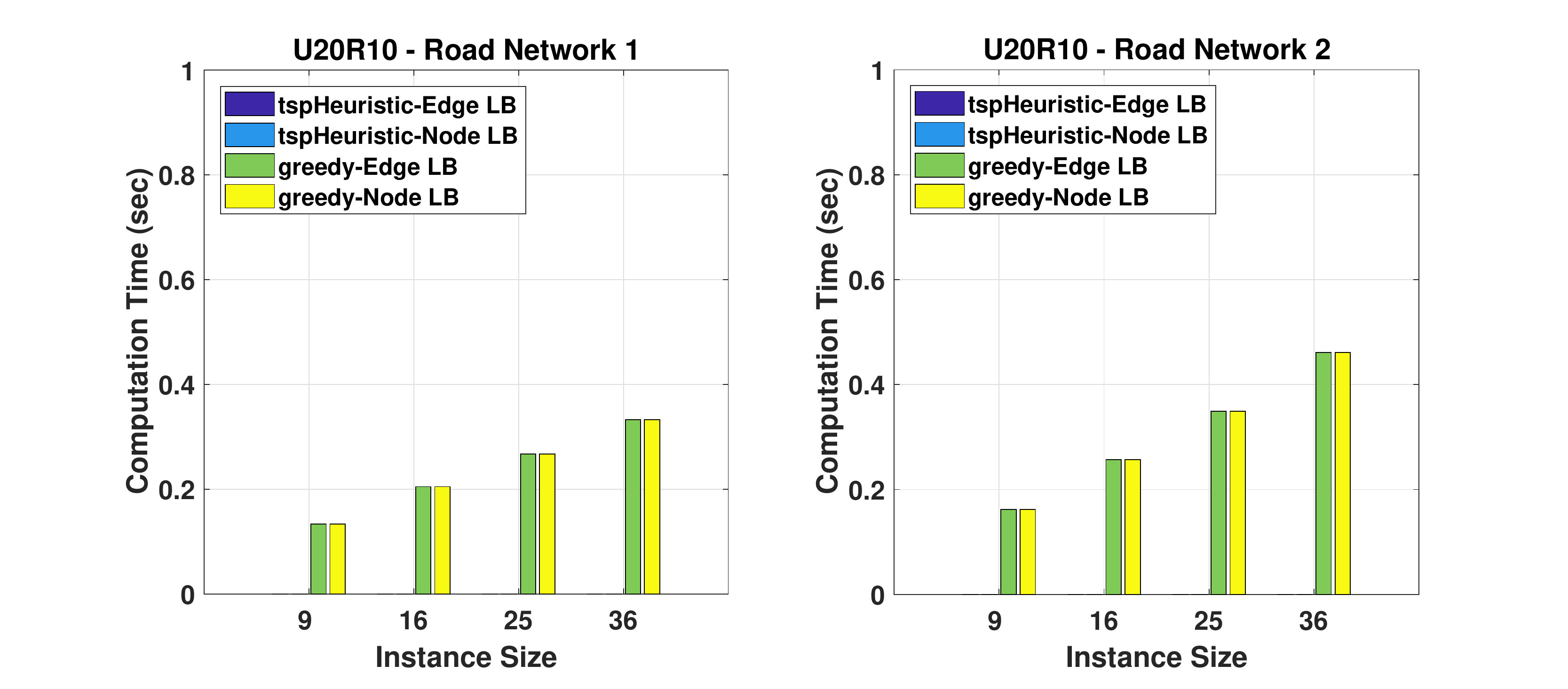}
	\caption{}\label{fig:simRoadNw2}	
\end{subfigure}
\begin{subfigure}{0.32\linewidth}
	\centering
	\includegraphics[clip,trim=20mm 5mm 0mm 5mm,width =  1.125\linewidth]{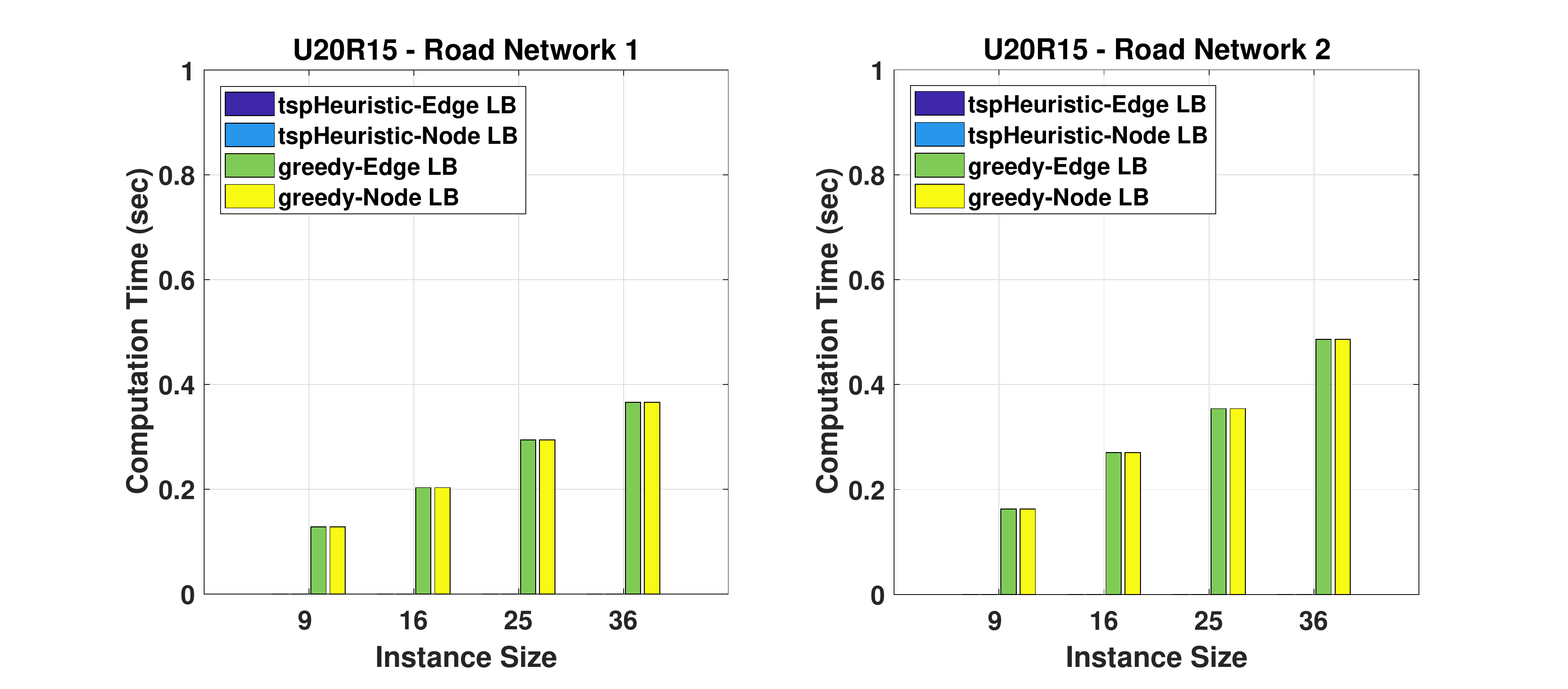}
	\caption{}\label{fig:simRoadNw1}	
\end{subfigure}
\begin{subfigure}{0.32\linewidth}
	\centering
	\includegraphics[clip,trim=20mm 5mm 0mm 5mm,width =  1.125\linewidth]{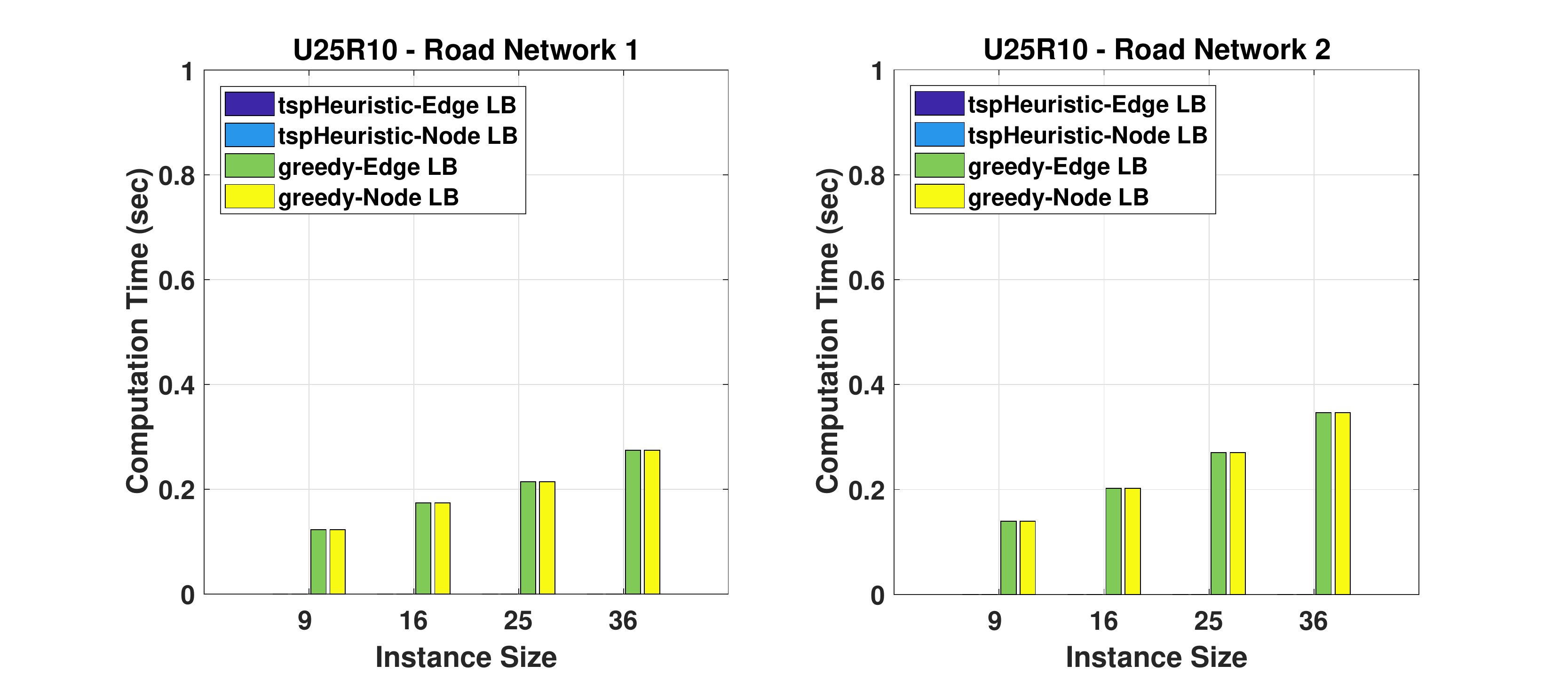}
	\caption{}\label{fig:simRoadNw2}	
\end{subfigure}
\begin{subfigure}{0.32\linewidth}
	\centering
	\includegraphics[clip,trim=20mm 5mm 0mm 5mm,width =  1.125\linewidth]{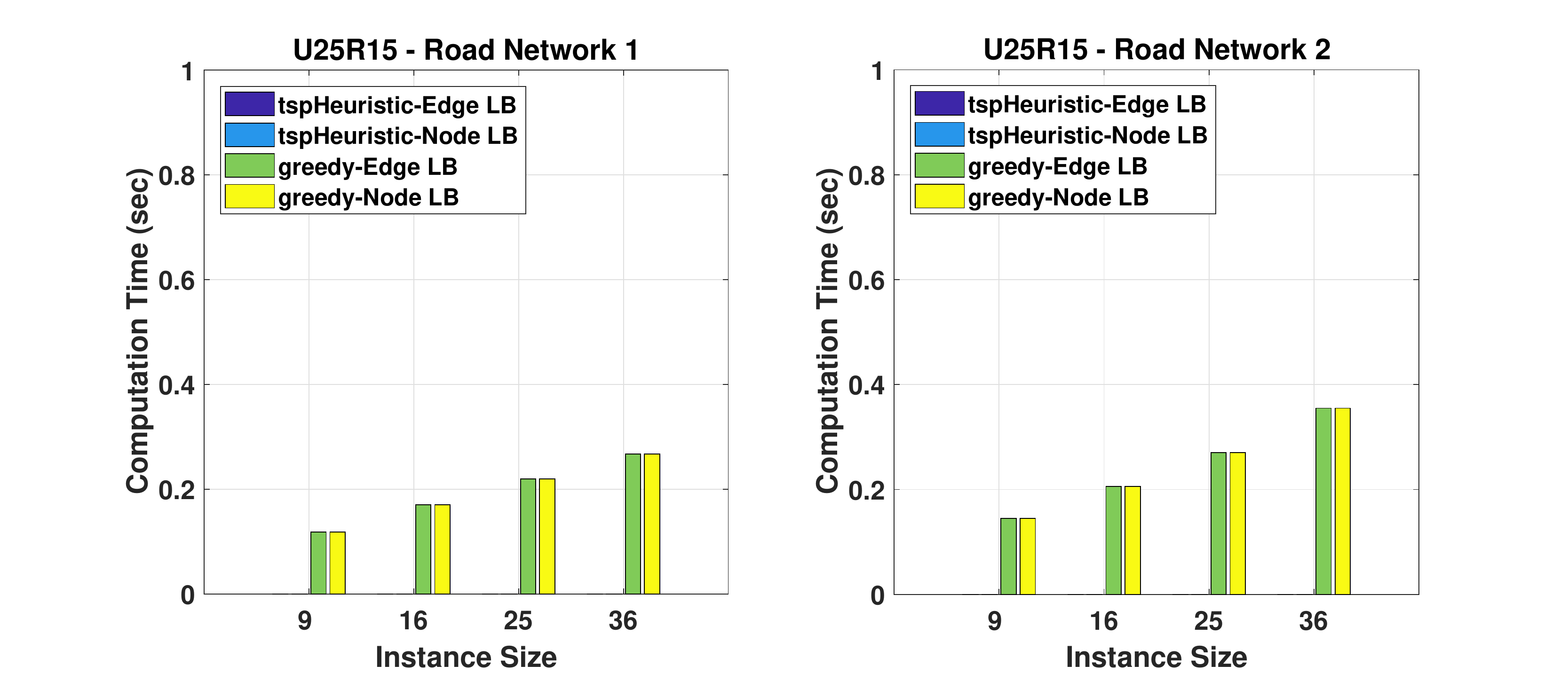}
	\caption{}\label{fig:simRoadNw2}	
\end{subfigure}
\caption{Time taken by the TSP-based heuristic and the greedy algorithm \cite{ICUASself} to compute feasible solutions.}\label{fig:timeHeuristic}
\end{figure*}
The TSP-based heuristic, however, does have it's own merits. As may be observed from Figure \ref{fig:timeHeuristic}, the heuristic is extremely fast and runs in microseconds. It consistency performs better than the greedy algorithm over all combinations of simulation parameters. This is very useful in missions that need computationally efficient solutions and can afford a higher mission cost. Figure \ref{fig:gapHeuristic} reports the relative gap for both the TSP-based heuristic and the greedy algorithm using lower bounds generated by the MILP formulations. Even though the mission cost is same, the relative gap is persistently lower for the edge based formulation. This admits the conclusion that the edge based formulation generates tighter lower bounds than the node based formulation. The results shown in Figure \ref{fig:gapMILP}, when inspected with this knowledge on lower bounds, may be understood with a different perspective. The node based formulation, does not necessarily generate worse solutions. The higher value of the relative gap is also attributed to relatively lose lower bounds.

\section{Conclusions and Future Directions}\label{sec:concFut}
This work addresses the cooperative routing problem for a fuel constrained UAV and a terrain constrained ground based refueling vehicle in the context of a coverage application. A two stage solution strategy is designed to find efficient solutions to the problem, wherein the first stage computes a feasible set of refueling sites for UAV-RV rendezvous and the second stage performs joint route planning that satisfies the fuel and speed limitations of the two vehicles, respectively. Alternate MILP formulations and a computationally efifcient heuristic algorithm are presented to solve the routing problem in the second stage. The construction-heuristic based on a TSP tour of the targets and a repair algorithm, allows for the development of evolutionary techniques as extensions to this work. A budgeted version of the problem that maximizes coverage with a cap on the number of refuels also makes an interesting variant of the problem. The literature on refueling problems also lacks any approximation schemes for the problem. Other interesting extensions to this problem, include multiple cost functions (for practitioners), pareto optimal solutions and multiple UAV scenarios.
\bibliographystyle{IEEEtran}
\bibliography{references}

\begin{thebibliography}{10}
\providecommand{\url}[1]{#1}
\csname url@rmstyle\endcsname
\providecommand{\newblock}{\relax}
\providecommand{\bibinfo}[2]{#2}
\providecommand\BIBentrySTDinterwordspacing{\spaceskip=0pt\relax}
\providecommand\BIBentryALTinterwordstretchfactor{4}
\providecommand\BIBentryALTinterwordspacing{\spaceskip=\fontdimen2\font plus
\BIBentryALTinterwordstretchfactor\fontdimen3\font minus
  \fontdimen4\font\relax}
\providecommand\BIBforeignlanguage[2]{{%
\expandafter\ifx\csname l@#1\endcsname\relax
\typeout{** WARNING: IEEEtran.bst: No hyphenation pattern has been}%
\typeout{** loaded for the language `#1'. Using the pattern for}%
\typeout{** the default language instead.}%
\else
\language=\csname l@#1\endcsname
\fi
#2}}

\bibitem{review2014uav}
I.~Colomina and P.~Molina, ``Unmanned aerial systems for photogrammetry and
  remote sensing: A review,'' \emph{ISPRS Journal of Photogrammetry and Remote
  Sensing}, vol.~92, pp. 79--97, 2014.

\bibitem{surveyForWildlife}
H.~Jachmann, \emph{Estimating abundance of African wildlife: an aid to adaptive
  management}.\hskip 1em plus 0.5em minus 0.4em\relax Springer Science \&
  Business Media, 2001.

\bibitem{sujit2007forestMonitoring}
P.~B. Sujit, D.~Kingston, and R.~Beard, ``Cooperative forest fire monitoring
  using multiple uavs,'' in \emph{IEEE Conference on Decision and Control}, Dec
  2007, pp. 4875--4880.

\bibitem{isr2005}
M.~Freed, W.~Fitzgerald, and R.~Harris, ``Intelligent autonomous surveillance
  of many targets with few uavs,'' in \emph{Proceedings of the Research and
  Development Partnering Conference, Department of Homeland Security, Boston,
  MA}, 2005.

\bibitem{physicalSurveysExpensive}
H.~Jachmann, ``Evaluation of four survey methods for estimating elephant
  densities,'' \emph{African Journal of Ecology}, vol.~29, no.~3, pp. 188--195,
  1991.

\bibitem{UAVsurvey2008}
J.~Everaerts, ``The use of unmanned aerial vehicles (uavs) for remote sensing
  and mapping,'' \emph{The International Archives of the Photogrammetry, Remote
  Sensing and Spatial Information Sciences}, vol.~37, pp. 1187--1192, 2008.

\bibitem{UAVsurvey2013BatteryProb}
\BIBentryALTinterwordspacing
C.~Vermeulen, P.~Lejeune, J.~Lisein, P.~Sawadogo, and P.~Bouché, ``Unmanned
  aerial survey of elephants,'' \emph{PLoS ONE}, vol.~8, no.~2, February 2013.
  [Online]. Available: \url{http://dx.doi.org/10.1371/Fjournal.pone.0054700}
\BIBentrySTDinterwordspacing

\bibitem{sundar2014TASE}
K.~Sundar and S.~Rathinam, ``Algorithms for routing an unmanned aerial vehicle
  in the presence of refueling depots,'' \emph{IEEE Transactions on Automation
  Science and Engineering}, vol.~11, no.~1, pp. 287--294, 2014.

\bibitem{kannon2014AFIT}
T.~E. Kannon, S.~G. Nurre, B.~J. Lunday, and R.~R. Hill, ``The aircraft routing
  problem with refueling,'' \emph{Optimization Letters}, pp. 1--16, 2014.

\bibitem{sundar2015formulations}
K.~Sundar, S.~Venkatachalam, and S.~Rathinam, ``Formulations and algorithms for
  the multiple depot, fuel-constrained, multiple vehicle routing problem,'' in
  \emph{American Control Conference}, July 2016, pp. 6489--6494.

\bibitem{khullerGasStation}
S.~Khuller, A.~Malekian, and J.~Mestre, ``To fill or not to fill: The gas
  station problem,'' \emph{ACM Transactions on Algorithms (TALG)}, vol.~7,
  no.~3, p.~36, 2011.

\bibitem{sundar2016exact}
K.~Sundar, S.~Venkatachalam, and S.~Rathinam, ``Analysis of mixed-integer
  linear programming formulations for a fuel-constrained multiple vehicle
  routing problem,'' \emph{Unmanned Systems}, 2017.

\bibitem{Levy2014}
D.~Levy, K.~Sundar, and S.~Rathinam, ``Heuristics for routing heterogeneous
  unmanned vehicles with fuel constraints,'' \emph{Mathematical Problems in
  Engineering}, vol. Article ID 131450, 2014.

\bibitem{erdougan2012green}
S.~Erdo{\u{g}}an and E.~Miller-Hooks, ``A green vehicle routing problem,''
  \emph{Transportation Research Part E: Logistics and Transportation Review},
  vol.~48, no.~1, pp. 100--114, 2012.

\bibitem{funke2015placement}
S.~Funke, A.~Nusser, and S.~Storandt, ``Placement of loading stations for
  electric vehicles: No detours necessary!'' \emph{Journal of Artificial
  Intelligence Research}, vol.~53, pp. 633--658, 2015.

\bibitem{smith2015TRO}
N.~Mathew, S.~Smith, and S.~Waslander, ``Multirobot rendezvous planning for
  recharging in persistent tasks,'' \emph{IEEE Transactions on Robotics},
  vol.~31, no.~1, pp. 128--142, Feb 2015.

\bibitem{ICUASself}
P.~Maini and P.~Sujit, ``On cooperation between a fuel constrained uav and a
  refueling ugv for large scale mapping applications,'' in \emph{International
  Conference on Unmanned Aircraft Systems}, June 2015, pp. 1370--1377.

\bibitem{2Echelon17}
Z.~Luo, Z.~Liu, and J.~Shi, ``A two-echelon cooperated routing problem for a
  ground vehicle and its carried unmanned aerial vehicle,'' \emph{Sensors},
  vol.~17, no.~5, p. 1144, 2017.

\bibitem{gautam2014ICUAS}
A.~Gautam, P.~Sujit, and S.~Saripalli, ``A survey of autonomous landing
  techniques for uavs,'' in \emph{Unmanned Aircraft Systems (ICUAS), 2014
  International Conference on}.\hskip 1em plus 0.5em minus 0.4em\relax IEEE,
  2014, pp. 1210--1218.

\bibitem{gautam2017IFAC}
\BIBentryALTinterwordspacing
------, ``Autonomous quadrotor landing using vision and pursuit guidance,''
  \emph{IFAC-PapersOnLine}, vol.~50, no.~1, pp. 10\,501 -- 10\,506, 2017, 20th
  IFAC World Congress. [Online]. Available:
  \url{http://www.sciencedirect.com/science/article/pii/S2405896317326204}
\BIBentrySTDinterwordspacing

\bibitem{dock1}
K.~A. Swieringa, C.~B. Hanson, J.~R. Richardson, J.~D. White, Z.~Hasan,
  E.~Qian, and A.~Girard, ``Autonomous battery swapping system for small-scale
  helicopters,'' in \emph{IEEE International Conference on Robotics and
  Automation}.\hskip 1em plus 0.5em minus 0.4em\relax IEEE, 2010, pp.
  3335--3340.

\bibitem{dock2}
K.~A. Suzuki, P.~Kemper~Filho, and J.~R. Morrison, ``Automatic battery
  replacement system for uavs: Analysis and design,'' \emph{Journal of
  Intelligent \& Robotic Systems}, vol.~65, no. 1-4, pp. 563--586, 2012.

\bibitem{Grotschel1985}
M.~Grotschel and M.~Padberg, ``Polyhedra theory,'' in \emph{The traveling
  salesman problem}, E.~Lawler, J.~Lenstra, A.~R. Kan, and D.~Shmoys,
  Eds.\hskip 1em plus 0.5em minus 0.4em\relax John Wiley \& Sons, New York,
  1985, pp. 251--305.

\bibitem{lkh}
S.~Lin and B.~W. Kernighan, ``An effective heuristic algorithm for the
  traveling-salesman problem,'' \emph{Operations research}, vol.~21, no.~2, pp.
  498--516, 1973.

\bibitem{Blyenburgh2007}
P.~van Blyenburgh, ``Unmanned aircraft systems: The global perspective,''
  \emph{UVS International, Paris, France}, 2007.

\end{thebibliography}

\end{document}